%% file: main.tex
\documentclass{article}
\usepackage{iclr2027_conference,times}
\input{math_commands.tex}

\usepackage{hyperref} 
\usepackage[capitalize,nameinlink,noabbrev]{cleveref}
\usepackage{url}
\usepackage{graphicx}
\usepackage{booktabs}
\usepackage{xcolor}
\usepackage{amsmath}
\usepackage{array}
\providecommand{\tightlist}{\setlength{\itemsep}{0pt}\setlength{\parskip}{0pt}}

\title{Cognitive Expert Language Models Better Align with the Corresponding Brain Systems}

\author{Zhivar Sourati\textsuperscript{1,3,*} \\
\texttt{souratih@usc.edu} \And
Mengxuan Helen Wu\textsuperscript{2,3,*} \\
\texttt{mengxuan@usc.edu} \And
Nona Ghazizadeh\textsuperscript{2,3} \\
\texttt{nghaziza@usc.edu} \AND
Jonas Kaplan\textsuperscript{2,3} \\
\texttt{jtkaplan@usc.edu} \And
Morteza Dehghani\textsuperscript{1,2,3} \\
\texttt{mdehghan@usc.edu} \And
Samuel A. Nastase\textsuperscript{2,3} \\
\texttt{snastase@usc.edu} \AND
\normalfont\textsuperscript{1}Department of Computer Science, University of Southern California \\
\textsuperscript{2}Department of Psychology, University of Southern California \\
\textsuperscript{3}Center for Computational Language Sciences, University of Southern California \\
\textsuperscript{*}Equal contribution}

\iclrfinalcopy 
\begin{document}

\maketitle
\lhead{}
\renewcommand{\headrulewidth}{0pt}

\begin{abstract}
Large language models (LLMs) can predict human brain activity across a variety of brain regions during natural language comprehension. Typically, however, LLM–brain alignment is measured using one model for different regions of the brain, and then model performance is summarized across regions. This one-model-fits-all approach ignores the functional specialization of brain regions. In this study, we assess whether a model oriented toward a particular cognitive domain aligns better with the brain system dedicated to that domain. Through prompting and fine-tuning, we first build ``expert LLM'' variants for six domains: sensory, spatial, numerical, reasoning, social, and abstract processing. We then examine whether each expert best predicts activity in the brain region associated with the corresponding cognitive domain. Consistent with our hypotheses, each expert's representations align more closely with the brain system most associated with the matching domain than do other experts. This holds under both prompting and fine-tuning, across three base models and three fMRI datasets. In a series of control analyses, we show that this model–brain alignment is specific to cognitive domain interventions; non-cognitive and surface-level interventions do not result in comparable alignment. Specializing models shifts regional alignment while leaving aggregate prediction accuracy largely unchanged, suggesting that summarizing alignment across regions may obscure regional differences in performance for specific models.
\end{abstract}

\section{Introduction}\label{sec:intro}

The human brain is not uniform in its functional architecture: it is topographically organized into partially specialized systems that support sensory processing, abstract meaning, spatial navigation, calculation, reasoning, and thinking about other people, among other functions. While these systems are not fully isolated modules, experimental work reveals a functional division of labor \citep{smith2009correspondence,kanwisher2010functional,amalric2016origins,dinicola2020parallel,petersen2024principles}. As large language models (LLMs) begin to reach, and sometimes surpass, human performance on many cognitive tasks \citep{webb2023emergent,alohali2025reasoning}, a growing body of work has begun to study the alignment between these models' internal representations and the brain's neural representations \citep{toneva2019interpreting,schrimpf2021neural,caucheteux2022brains,goldstein2022shared,kumar2024shared}. Model–brain alignment is often measured using an encoding model. Participants view or listen to a set of stimuli while their brain activity is recorded, and the same stimuli are also passed through the model to obtain its internal representations. A linear mapping is then fit to predict brain activity from these features.  A model that yields strong predictions for a given brain region (typically relative to competing models) is interpreted to represent features of the stimulus encoded in that region 
\citep{naselaris2011encoding,dupre2025voxelwise}. While LLM features provide strong predictions across many brain areas, they are often tested under a one-model-fits-all approach: a single set of features is fit across all regions and performance is summarized across regions \citep{schrimpf2021neural}, collapsing across functionally distinct brain systems.

\begin{figure}[t]
\centering
\includegraphics[width=\linewidth]{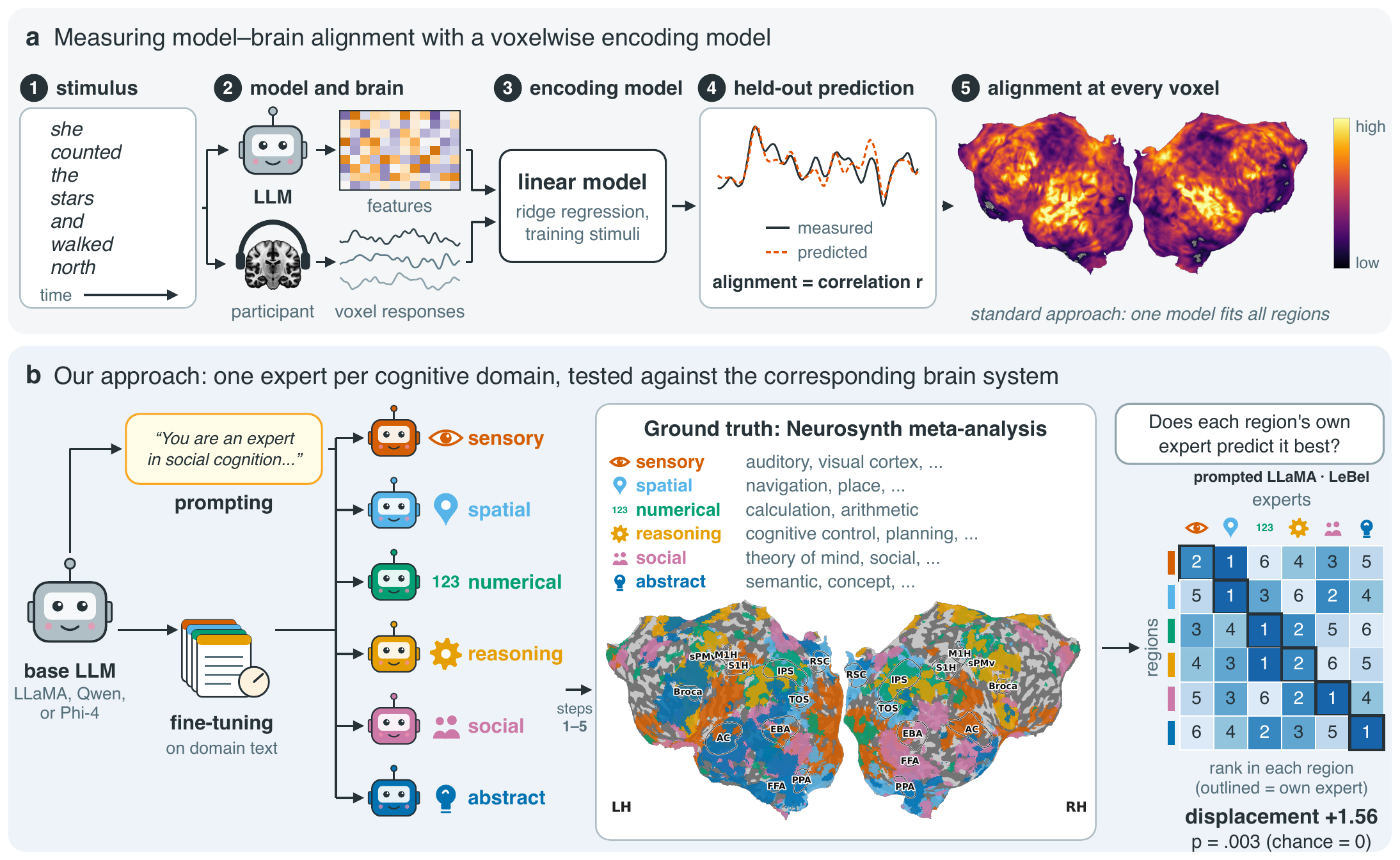}
\caption{\textbf{Measuring alignment with specialized brain systems.} \emph{(a)} A voxelwise encoding model predicts each voxel's response to a stimulus from a language model's features and is scored on held-out data, giving an alignment map over the cortex (here for the base model, in one LeBel participant). \emph{(b)} We build one expert per cognitive domain by prompting or fine-tuning, pass each through the same steps, and ask whether each domain's Neurosynth region is predicted best by its own expert; the matrix ranks the six experts in each region for prompted LLaMA on the LeBel dataset, the running example of \Cref{fig:lead} (displacement is computed per participant, then averaged).}
\label{fig:overview}
\end{figure}

The success of neural network models in predicting brain activity is thought to reflect a pressure to converge on similar internal representations when optimized to perform sufficiently complex, human-like tasks \citep{yamins2026contravariance}. However, the one-model-fits-all approach is in tension with the functional division of labor observed in the human brain, where different brain systems play a larger role in some tasks than in others. Domain specialization is becoming commonplace in LLMs as well: models are narrowed to excel in one kind of content (i.e., experts), most commonly by fine-tuning their weights on domain data \citep{li2022branch,fu2023specializing,biderman2024lora} or by prompting them with a specialized role \citep{li2023camel,xu2023expertprompting,kong2024better}. Specialization is usually evaluated according to the model's behavior: its performance on in-domain tasks and other behavioral cues \citep{fu2023specializing,wang2025mixture,zheng2024helpful}. Behavioral evaluation leaves open the question of how specialization changes a model's internal structure. The brain provides a useful reference for comparison, with its own division of labor across cognitive domains. By evaluating whether an expert model aligns with specialized brain systems, we can learn whether and how a model's internal representations have been reshaped.

In the current study, we investigate whether specializing a model toward a particular cognitive domain improves model-brain alignment in a manner consistent with the brain's topographic functional specialization. We introduce six ``expert'' variants of an LLM, each specialized for one cognitive domain through one of two interventions, prompting or fine-tuning: sensory, spatial, numerical, reasoning, social, or abstract. We then test whether each expert better predicts brain activity than other experts in the corresponding brain regions, defined using the meta-analytic database Neurosynth \citep{yarkoni2011large} (\Cref{fig:overview}). Three key findings support this hypothesis: (1) Each expert aligns significantly more with the brain system for the matching domain than the other experts, and this pattern holds across interventions, base models, outcome metrics, and datasets involving continuous narratives or isolated sentences. (2) This regional alignment does not emerge after control interventions that alter the model without improving cognitive expertise, such as stylistic fine-tuning or arbitrary system prompts. (3) Moreover, model specialization improves alignment with certain brain systems while leaving overall alignment largely unchanged, suggesting that studies focusing on global alignment may overlook representational reorganization. These findings speak to a central question about LLMs: whether models trained to simply predict text also learn latent representations of other cognitive domains \citep{mahowald2023dissociating,fedorenko2024language}, and whether representations can be surfaced or enhanced through post-training interventions. We find that specializing a model toward, e.g., social or numerical content, even by prompting alone, shifts its representations toward the brain regions for those functions, which lie largely outside the language network. Text-trained models may thus already carry representations of these functions, and specialization can bring them to the fore. For neuroscience, specialized LLMs can serve as manipulable model systems for in-silico experiments on specific brain systems \citep{jain2024computational,d2026foundation}.

\section{Related Work}\label{sec:related}

Studies comparing LLMs with the brain have varied model architecture, objective, and inputs in hopes of identifying model features that drive model–brain alignment \citep{toneva2019interpreting,schrimpf2021neural,caucheteux2022brains,goldstein2022shared,antonello2023scaling,kumar2024shared,singh2025evaluating}. These studies typically converge on a single best performing model across many brain areas, suggesting that a simple linear re-weighting of LLM features (learned per voxel/area by the encoding model) is sufficient to capture brain activity across areas. This one-model-fits-all approach obscures the possibility that models tuned (or prompted) for different tasks may yield internal representations that better align with certain brain systems in a way that cannot be reduced to linear re-weighting of model features. 

A complementary line of work in this vein has suggested that fine-tuning on certain natural language tasks, such as sentiment analysis, question answering, or coreference resolution, may improve alignment over base models for listening or reading tasks \citep{oota2022neural}. Reports on the benefits of specializing a pretrained model for certain tasks have been mixed, however. For example, another group found that fine-tuning or prompt-tuning on ten language-understanding tasks did not improve model–brain alignment over the base model for most tasks \citep{sun2023fine}. \citet{aw2024instruction} argued, across three fMRI datasets, that instruction-tuned models and models trained on conversations do in fact improve model–brain alignment over base models in the human language network. On the other hand, \citet{gao2025increasing} found that instruction-tuned LLaMA models (Alpaca and Vicuna) did not outperform base models in predicting fMRI activity during naturalistic reading. Finally, recent work by \citet{oota2025task} using instruction-tuned multimodal LLMs with a movie-watching fMRI dataset found that prompted models yield stronger model–brain alignment than base models. Interestingly, in interpreting their findings, \citet{oota2025task} suggest that certain task instructions may enhance alignment more in certain regions than others; for example, a ``narrative understanding'' prompt yields the largest boost in angular gyrus, an area typically associated with high-level semantic processing. The link between model expertise and functional specialization remains unclear, however, because the regions under consideration were not defined by their functions and their functions do not generally correspond to the instruction-tuning tasks. This marks the jumping-off point for the present paper: we specialize models toward six widely studied cognitive domains and compare their alignment with meta-analytically defined brain systems matched to the same domains. Through targeted model comparisons, we ask how different paths to specialization reshape a model's representations, with the brain's functional division of labor as the reference.

\section{Methods}\label{sec:methods}

\subsection{Constructing and Validating the Expert Models}\label{sec:experts}

Throughout the paper, an \textbf{expert} is a variant of a base model specialized for one cognitive domain and a \textbf{family} is the set of six experts derived from one base model under one intervention. We build experts in two ways: by \textbf{prompting}, in which a domain instruction is placed in the input and the weights remain untouched, and by \textbf{fine-tuning}, in which LoRA adapters \citep{hu2021lora} are trained on domain-specific text. Both interventions are applied to three base models (LLaMA-3-8B; \citealp{grattafiori2024llama}, Qwen3-8B; \citealp{yang2025qwen3}, and Phi-4-14B; \citealp{abdin2024phi}), resulting in one prompted and one fine-tuned family per base model, each with an expert for the sensory, spatial, numerical, reasoning, social, and abstract-concept (``abstract'' for short) domains. Further details are given in \Crefrange{app:experts}{app:inference}; \Cref{fig:overview}b summarizes the design.

\paragraph{Prompting.} Each prompted expert receives a fixed system prompt, the same for every dataset, that names its domain and asks the model to focus on it (\citealp{xu2023expertprompting}; \Cref{app:experts}).

\paragraph{Fine-tuning.} Each fine-tuned expert is a LoRA adapter trained on instruction-response pairs from its own domain, including instruction-following corpora, curated text, and distilled examples. Checkpoints are selected under a guard that keeps general-text perplexity close to the base model's, so specialization does not cost general language ability (\citealp{luo2025empirical}; \Cref{app:experts}).

\paragraph{Validation.} Each expert must first behaviorally demonstrate that it is specialized for its domain. Fine-tuned experts are assessed by perplexity on held-out domain text: we expect each expert to improve most on the matching domain text among the six experts (six-way ranking). Prompted experts are assessed on what they generate: continuations of neutral story openings are scored against domain lexicons from which every instruction word is removed, and are rated by a separate LLM judge \citep{zheng2023judging}, blind to the prompt, for how strongly they engage each domain; a six-domain benchmark battery (MMLU; \citealp{hendrycks2020measuring}) serves as a further check (\Cref{app:behavior}). 

\paragraph{Control models.} To test whether alignment is specific to cognitive content, we build five control families with the same construction procedures but no cognitive content: \textbf{random-LoRA}, untrained adapters matched in norm to the cognitive ones; \textbf{seed-only}, six adapters trained on one pooled dataset that mixes all six domains, differing only in sampling and initialization; \textbf{random-prompt}, system prompts that make the model an expert in fields such as zebra striping anatomy or vintage typewriter restoration; \textbf{surface-form}, adapters trained on style-transformed text such as leetspeak and pirate English; and \textbf{keyword}, each domain's bare vocabulary with no instruction (\Cref{app:brain}).

\subsection{Measuring Model–Brain Alignment}\label{sec:brain}

Model–brain alignment is measured using a linear encoding model \citep{dupre2025voxelwise}. Participants listen to or read language during fMRI, and the same language is passed through the model to obtain its internal representations. A linear mapping from these representations to human neural activity is estimated from a subset of training data, then evaluated on held-out test data. We apply this procedure to every expert on three fMRI datasets.

\paragraph{Datasets and brain regions.} We evaluate brain alignment on three fMRI datasets: three densely sampled subjects listening to spoken narratives (``LeBel''; \citealp{lebel2023natural}), the 49-subject English cohort of Le Petit Prince (``LPP''; \citealp{li2022petit}), and ten subjects reading passages one sentence at a time (``Pereira''; \citealp{pereira2018toward}). The first two datasets comprise naturalistic narratives, offering greater ecological validity \citep{hamilton2020revolution,nastase2020keep} and engaging varying cognitive/affective processing across many regions of the brain \citep{baldassano2017discovering,horowitz2026engagement,vaccaro2024neural,wu2025first}. This lets us measure domain-specific alignment as it is expressed within the same naturalistic comprehension task, reducing potential biases introduced by engineering different tasks for different domains. The third dataset presents isolated sentences and tests whether the effect survives under a more controlled stimulus format. 

Brain regions for the six domains were obtained from Neurosynth \citep{yarkoni2011large}, an automated meta-analysis of the fMRI literature. For any term that is widely used in published studies, Neurosynth can generate a whole-brain map of how reliably each voxel is reported active in studies mentioning that term. For each domain we combine the maps of a small set of related terms and threshold the result into a binary mask on a common template, a network of several areas that need not be contiguous, which we call the domain's \emph{region} for brevity. These are projected into each subject's functional space and are shown on the cortical surface in \Cref{fig:overview}b (all three LeBel subjects in \Cref{fig:regions}). The Pereira dataset includes a parcellation into five functional networks: language \citep{fedorenko2011functional}, multiple-demand \citep{fedorenko2013broad}, default-mode, visual, and auditory \citep{power2011functional}. For simplicity, we assign each network to one domain and use it in place of the Neurosynth regions. The term lists, thresholds and projection are detailed in \Cref{app:brain}.

\paragraph{Encoding models.} Following the pipeline of \citet{singh2025evaluating}, each language stimulus is embedded as a sequence of 10-word contexts, with each family receiving the text in the corresponding form (prompted experts with their system prompt, fine-tuned experts without one). Features are reduced by PCA retaining 90\% of variance, expanded with eight finite-impulse-response delays to accommodate the hemodynamic lag \citep{huth2012continuous}, and regressed onto the fMRI activity. The fMRI activity is reduced to 100 principal components for model fitting and projected back into voxel space for model evaluation. We observe qualitatively similar results when fitting voxelwise models (\Cref{app:brain}). Models are estimated using ridge regression with a penalty selected by bootstrap cross-validation within the training stories. Train and test stories follow each dataset's standard split reported in the corresponding papers. Encoding accuracy is quantified as the Pearson correlation between model-predicted and actual responses on the held-out test stories (\Cref{fig:overview}a). The adaptations of this pipeline for LPP and Pereira are given in \Cref{app:brain}.

\paragraph{Regional alignment metrics.} For each brain system, we rank the six experts of a family by how well each predicts activity inside that region relative to outside it: each expert's encoding accuracy, with the family mean across experts subtracted at every voxel, is averaged over the voxels inside the region and over those outside it, and the six experts are ranked on the difference (\Cref{fig:overview}b, right). Our primary statistic, \textbf{displacement}, takes the rank of each region's own expert (1 = best), averages it over regions, and subtracts it from the chance rank of 3.5, the midpoint of 1 to 6, so 0 is chance and positive values mean that matched experts rank nearer the top; because six experts can be assigned to six regions in exactly 720 ways, its null distribution is exact \citep{nichols2002nonparametric}, and $p$ is the fraction of relabelings scoring at least as high (120 relabelings and a chance rank of 3.0 on Pereira's five networks). Alongside it we report \textbf{$\Delta r$}, the same comparison in correlation units: the expert's mean encoding accuracy inside the matching region minus that of the other five experts, averaged over regions and tested across subjects. Both compare each expert with the other five: displacement keeps only the rank of its advantage inside the region over outside it, whereas $\Delta r$ keeps the size of its advantage inside the region. Hyperparameters such as the readout layer and the adapter's rank were set by a sweep, and the effect holds with this selection included: put through the same sweep, relabeled experts reach a mean displacement of +0.31, against +0.55 for the actual experts ($p$ = .0002; \Cref{app:inference}).

\paragraph{Evaluating the controls.} The control families are passed through the same encoding pipeline. Because a control family has no true expert-to-region mapping, scoring it under every possible assignment of its members to the regions yields a null distribution of displacement, centered on zero. We then ask whether the expert families exceed that distribution, comparing each cognitive result with a content-free control and with the control that shares its intervention.

\section{Results}\label{sec:results}

\subsection{Behavioral Validation of the Expert Models}\label{sec:behavior}

\begin{figure}[t]
\centering
\includegraphics[width=0.9\linewidth]{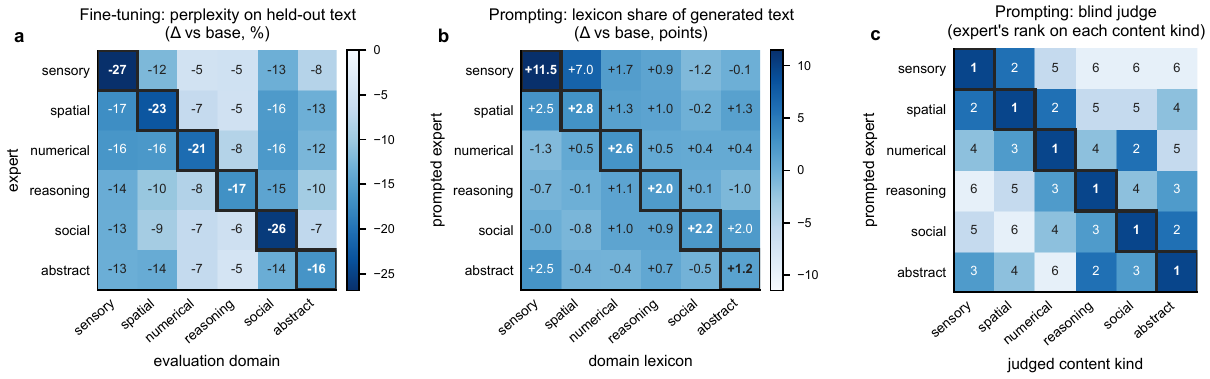}
\caption{\textbf{Behavioral validation of the experts.} LLaMA. \emph{(a)} Fine-tuning: change in perplexity (\%) relative to the base model, each expert (row) on each domain's held-out text (column). \emph{(b)} Prompting: change in the share of generated words from each domain's lexicon (percentage points). \emph{(c)} Prompting: each expert's rank on each content kind by a blind judge (1 = highest). Outlined cells mark the matched expert; each column compares the six experts on one domain.}
\label{fig:behavior}
\end{figure}

Before incorporating brain data, we first test whether each expert is behaviorally specialized for its respective domain. For the fine-tuned experts, we measure perplexity on held-out text from each domain. In every fine-tuned family, each expert's perplexity is improved most for its own domain text among the six experts, with mean own-domain improvements of 22\% to 40\% over the base model (\Cref{fig:behavior}a; all nine families in \Cref{app:behavior}). The prompted experts are specialized only while the instruction is present, so we measure the content of what they generate under this prompt. Each expert continues the same neutral story openings, and we score the continuations in two ways. In the first approach, we quantify the share of generated words that belong to each domain's lexicon, with the instruction words removed from the lexicons so that an expert cannot score by echoing its prompt. In the second approach, a separate model, blind to the prompt, rates how strongly each continuation engages each domain, after the instructions are removed. On both measures, the matched prompting expert outperforms the base model across all three model families. By the lexicon measure, the expert produces 3.7, 1.9, and 3.1 more words per 100 from its own domain than the base model does, for LLaMA, Qwen, and Phi-4 respectively (exact permutation $p$ = .001, .043, and .001). By the judge measure, an expert's own domain gains 1.70, 0.24, and 1.41 more points (on a 0--10 scale) relative to the base model than other domains do ($p$ = .001, .018, and .001). Figure~\ref{fig:behavior}b (lexicon) and c (judge) show results for LLaMA. Results for the other base models, a second judge, and a performance-based evaluation are reported in \Cref{app:behavior}.

\subsection{Experts Improve Alignment in Matched Brain Regions}\label{sec:localization}

\begin{figure}[t]
\centering
\includegraphics[width=0.92\linewidth]{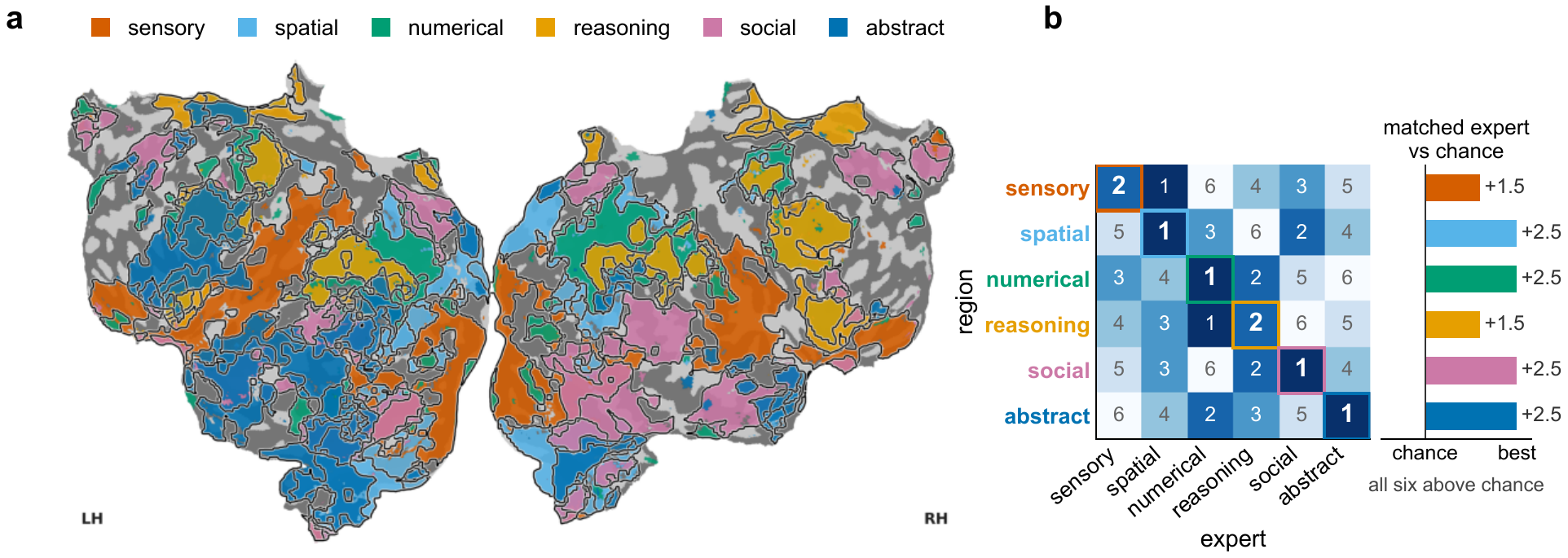}
\caption{\textbf{Expert models are aligned with the corresponding brain systems.} Prompted LLaMA on LeBel. \emph{(a)} The six regions on the exemplar subject's cortical surface, each filled with its own expert's color where that expert ranks above chance (the other two subjects are in \Cref{fig:lead_subjects}). \emph{(b)} Each expert's rank within each region (row; 1 = best, chance = 3.5) on its accuracy inside minus outside the region relative to the family mean, averaged over subjects (\Cref{tab:main} averages per-subject displacement instead). Outlined cells are the matched experts; bars give their rank relative to chance.}
\label{fig:lead}
\end{figure}

We next ask whether each expert best predicts activity in the brain region associated with its own domain. We find that in every one of the eighteen combinations of base model, intervention and dataset, the matched experts rank above chance in their own regions, giving a positive displacement (\Cref{tab:main}), and twelve of the eighteen are individually significant under the exact permutation test, spread across both interventions and all three datasets, and in most regions the matched expert ranks above chance (Regions in \Cref{tab:main}). \Cref{fig:lead} shows the running example, prompted LLaMA on the LeBel dataset: every region's own expert ranks first or second in that region (displacement +1.56, p = .003), and the per-expert cortical maps behind those ranks are shown in \Cref{app:alignment}. The same results hold when the ranks are replaced by the continuous contrast: the matched expert's encoding accuracy inside its region minus the other experts' ($\Delta r$ in \Cref{tab:main}) is positive in all eighteen combinations and significant in twelve. We observe the same pattern under other ways of scoring regional alignment, including restricting the comparison to best-predicted voxels and reading the Neurosynth maps as continuous gradients rather than binary regions (\Cref{app:alignment}).

Both interventions produce a positive effect on every base model. Combining the 62 subjects of the three datasets for each base model and intervention, the effect is significant for LLaMA prompting ($p$ \textless{} .001), LLaMA fine-tuning ($p$ = .002), Qwen fine-tuning ($p$ = .033), and Phi-4 fine-tuning ($p$ \textless{} .001), and nearly so for Qwen prompting ($p$ = .051) and Phi-4 prompting ($p$ = .058). Each subject's displacement is standardized by its own permutation null before combining (\Cref{app:inference}). Additionally, behavioral specialization and alignment are related across base models. Under prompting, LLaMA's experts are the most behaviorally specialized in what they generate, followed by Phi-4's and then Qwen's (judge advantage +1.70, +1.41, and +0.24; \Cref{sec:behavior}), and this is also the order of their mean displacement over the three datasets (+0.84, +0.58, and +0.46). Under fine-tuning, ranking base models by how much more each expert improves its own domain's text than the other experts do also gives their displacement order (\Cref{app:behavior}). This agreement across the two rankings is significant ($p$ = .03; the probability that two rankings would agree by chance, computed by randomly shuffling the base models within each intervention). Within a family, however, ranking a base model's six experts by specialization does not recover their ranking by alignment (\Cref{app:interventions}).

\begin{table}[t]
\caption{\textbf{Regional alignment for every model family, intervention, and dataset.} Displacement: 3.5 minus the mean rank of each region's own expert (3.0 on Pereira; 0 = chance), with its exact relabeling $p$. vs null: percentile within the seed-only control family's displacement distribution at the same setting. $\Delta r$: the matched expert's accuracy inside its region minus the other experts', tested across subjects. Regions: regions whose own expert ranks above chance. Bold: $p$ \textless{} .05 (vs null: beyond the 95th percentile). Pooled rows give the mean over all 62 subjects, with $p$ from a test that standardizes each subject's displacement by its own permutation null (\Cref{app:inference}).}
\label{tab:main}
\centering
\scriptsize
\setlength{\tabcolsep}{3.5pt}
\renewcommand{\arraystretch}{0.92}
\begin{tabular}{lllrrrrrrr}
\toprule
Model & Intervention & Dataset & $n$ & Displacement & $p$ & vs null & $\Delta r \times 10^{3}$ & $p$ & Regions \\
\midrule
LLaMA & Prompting & LeBel & 3 & \textbf{+1.56} & \textbf{.003} & \textbf{100th} & \textbf{+1.25} & \textbf{.011} & 6/6 \\
 &  & LPP & 49 & \textbf{+0.42} & \textbf{.017} & \textbf{100th} & \textbf{+0.39} & \textbf{$<$.001} & 5/6 \\
 &  & Pereira & 10 & \textbf{+0.54} & \textbf{.017} & \textbf{100th} & \textbf{+6.51} & \textbf{.013} & 4/5 \\
  &  & \emph{all pooled} & 62 & \textbf{+0.49} & \textbf{$<$.001} &  & \textbf{+1.42} & \textbf{.002} &  \\
\addlinespace[2pt]
 & Fine-tuning & LeBel & 3 & \textbf{+0.89} & \textbf{.019} & \textbf{99th} & +0.77 & .117 & 3/6 \\
 &  & LPP & 49 & \textbf{+0.30} & \textbf{.014} & \textbf{98th} & \textbf{+0.41} & \textbf{$<$.001} & 4/6 \\
 &  & Pereira & 10 & +0.34 & .133 & \textbf{95th} & \textbf{+1.36} & \textbf{.037} & 3/5 \\
  &  & \emph{all pooled} & 62 & \textbf{+0.34} & \textbf{.002} &  & \textbf{+0.58} & \textbf{$<$.001} &  \\
\addlinespace[2pt]
Qwen & Prompting & LeBel & 3 & \textbf{+0.89} & \textbf{.007} & \textbf{98th} & \textbf{+0.87} & \textbf{.048} & 5/6 \\
 &  & LPP & 49 & +0.07 & .321 & 61st & \textbf{+0.15} & \textbf{.041} & 4/6 \\
 &  & Pereira & 10 & \textbf{+0.43} & \textbf{.042} & \textbf{98th} & \textbf{+3.30} & \textbf{.029} & 3/5 \\
  &  & \emph{all pooled} & 62 & +0.17 & .051 &  & \textbf{+0.69} & \textbf{.009} &  \\
\addlinespace[2pt]
 & Fine-tuning & LeBel & 3 & \textbf{+0.61} & \textbf{.017} & \textbf{97th} & +0.85 & .084 & 5/6 \\
 &  & LPP & 49 & +0.42 & .065 & \textbf{100th} & \textbf{+0.82} & \textbf{$<$.001} & 4/6 \\
 &  & Pereira & 10 & +0.08 & .400 & 65th & +0.59 & .319 & 3/5 \\
  &  & \emph{all pooled} & 62 & \textbf{+0.37} & \textbf{.033} &  & \textbf{+0.78} & \textbf{$<$.001} &  \\
\addlinespace[2pt]
Phi-4 & Prompting & LeBel & 3 & \textbf{+1.17} & \textbf{.003} & \textbf{99th} & +0.84 & .118 & 6/6 \\
 &  & LPP & 49 & +0.17 & .214 & 92nd & \textbf{+0.18} & \textbf{.034} & 3/6 \\
 &  & Pereira & 10 & \textbf{+0.42} & \textbf{.033} & \textbf{99th} & +3.46 & .123 & 3/5 \\
  &  & \emph{all pooled} & 62 & +0.26 & .058 &  & +0.74 & .059 &  \\
\addlinespace[2pt]
 & Fine-tuning & LeBel & 3 & \textbf{+0.94} & \textbf{.007} & \textbf{100th} & \textbf{+0.65} & \textbf{.039} & 4/6 \\
 &  & LPP & 49 & \textbf{+0.33} & \textbf{.008} & \textbf{99th} & \textbf{+0.45} & \textbf{$<$.001} & 5/6 \\
 &  & Pereira & 10 & +0.34 & .092 & 85th & +2.59 & .053 & 5/5 \\
  &  & \emph{all pooled} & 62 & \textbf{+0.36} & \textbf{$<$.001} &  & \textbf{+0.81} & \textbf{.001} &  \\

\bottomrule
\end{tabular}
\end{table}

\subsection{The Effect of Prompting versus Fine-tuning on Regional Alignment}\label{sec:interventions}

\begin{figure}[t]
\centering
\includegraphics[width=0.92\linewidth]{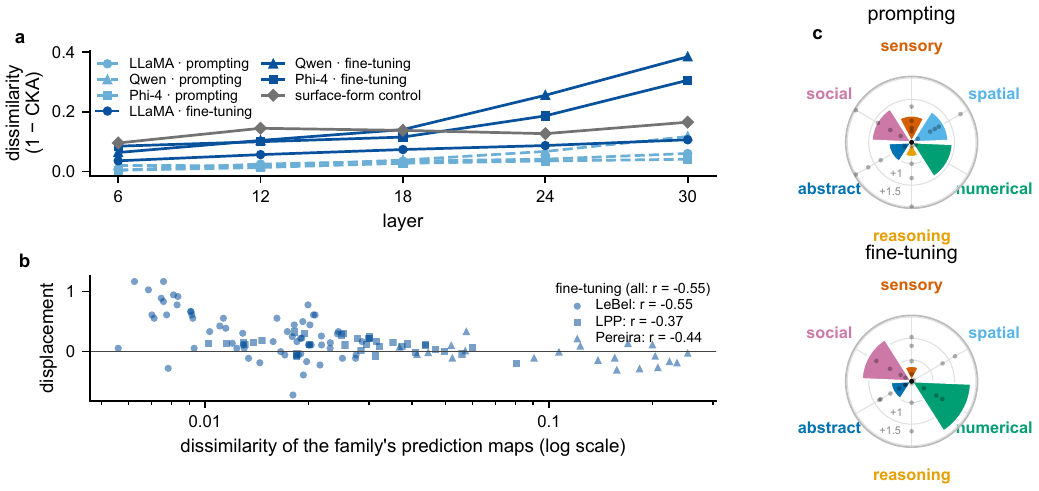}
\caption{\textbf{Comparing the two interventions.} \emph{(a)} Dissimilarity of the six experts' stimulus representations (1 $-$ mean pairwise CKA) by layer, for every family on LPP and for the surface-form control. \emph{(b)} Dissimilarity of each fine-tuned family's six prediction maps (1 $-$ mean pairwise correlation) against its displacement, one point per model, dataset and layer. \emph{(c)} Each expert's own-region displacement under each intervention (rings 0.5 apart), averaged over the narrative datasets and base models; dots are single configurations, and values below chance sit at the center.}
\label{fig:interventions}
\end{figure}

Prompting and fine-tuning change a model differently, so we asked whether the experts they produce align with the brain differently. Across the two narrative datasets and three base models, we compared the interventions on several aspects of model--brain alignment: its strength, its consistency across subjects, the layer at which it peaks, and its sensitivity to the layer, adapter rank, region definition, and scoring statistic (\Cref{app:interventions}). Neither intervention is consistently ahead on any of these; which one leads depends on the base model or dataset (\Cref{fig:route_ledger}). The one property that does separate them is the scale of representational change: by centered kernel alignment (CKA; \citealp{kornblith2019similarity}), fine-tuning moves the six experts' representations 1.4 to 6.8 times further apart than prompting does, in all six base model--dataset combinations, and increasingly with depth (\Cref{fig:interventions}a). Since fine-tuning pulls experts apart more, we asked whether that separation improves alignment to brain systems. We find that it does not: across the fine-tuned families, greater divergence among a family's six prediction maps corresponds to poorer alignment among its experts ($r=-0.55$ overall; $-0.55$, $-0.37$, and $-0.44$ within LeBel, LPP, and Pereira; \Cref{fig:interventions}b). A larger perturbation of the model thus does not necessarily yield better alignment (see also \Cref{sec:specificity}).

The interventions also differ in which experts align with their respective regions (\Cref{fig:interventions}c). Prompting aligns the spatial expert in all six combinations of base model and narrative dataset (mean own-region displacement +0.86), whereas fine-tuning's spatial expert sits at chance on average (0.00). The difference is reliable: for each subject in the LPP dataset, prompting's spatial expert outperforms fine-tuning's (pooled t(48) = 5.8, p \textless{} .0001), for each base model and at every layer, adapter rank and region definition. The numerical expert shows the reverse: for each subject in the LPP dataset, experts produced by fine-tuning outperform prompting at every layer (t(48) between 2.3 and 5.2, all p \textless{} .03). Only the brain data reveal the spatial difference; in behavior the spatial expert is the second or third most specialized by lexicon share and perplexity in every base model (\Cref{app:interventions}). The two interventions thus reach comparable alignment through different experts.

\subsection{Specificity to Cognitive Content}\label{sec:specificity}

\begin{figure}[t]
\centering
\includegraphics[width=\linewidth]{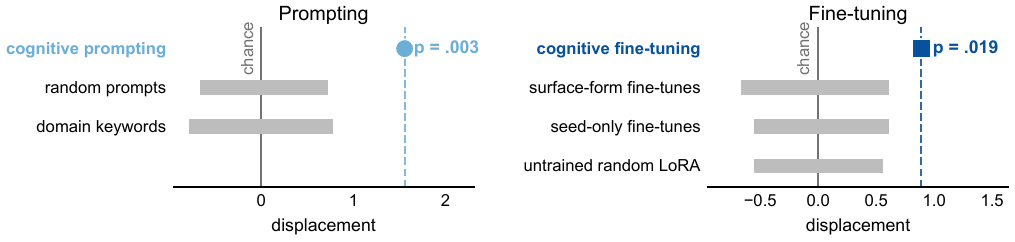}
\caption{\textbf{Cognitive domain experts yield stronger regional alignment than control interventions.} LLaMA on LeBel, each intervention at its \Cref{tab:main} setting. Bands: the central 90\% of a control family's displacement over all 720 assignments of its members to the regions. Points: the expert family, each expert paired with its own domain's region, with exact permutation $p$.}
\label{fig:route_controls}
\end{figure}

Regional alignment could, in principle, follow from any change to a model, whatever its content. To test this we use the control families of \Cref{sec:experts}, built by the same procedures but with no cognitive content. Each control family is scored under all 720 assignments of its members to the regions (\Cref{sec:brain}), and each expert family is placed within that null distribution. For the running example's base model (LLaMA), both cognitive interventions lie beyond the 95th percentile of every corresponding control distribution (\Cref{fig:route_controls}); over all eighteen combinations of base model, intervention and dataset, the expert families stand on average at the 94th percentile of the null from the six fine-tunes trained on the same pooled data, whose members differ in no content, and fourteen of them lie beyond its 95th percentile (\Cref{app:controls}). Alignment is also not a matter of how much an intervention changes the model. Displacement asks where the differences between a family's members fall; a separate measure, the size of the change, asks only how large they are (\Cref{app:controls}). By that measure, the surface-form fine-tunes differ nearly four times as much as the cognitive experts do, yet only the cognitive experts' differences fall in the regions their domains predict.

As a stronger control, the keyword family gives each domain's vocabulary without instructing the model to act as an expert in it. The cognitive experts yielded stronger regional alignment than their keyword-based counterparts in seven of nine base model $\times$ dataset comparisons; the exceptions are Qwen on the two narrative datasets, where vocabulary alone matches or exceeds the instruction, consistent with Qwen's prompted experts being the least differentiated behaviorally (\Cref{tab:keyword}).

\subsection{The Effect of Specialization on Overall Prediction Accuracy}\label{sec:dissociation}

\begin{figure}[t]
\centering
\includegraphics[width=\linewidth]{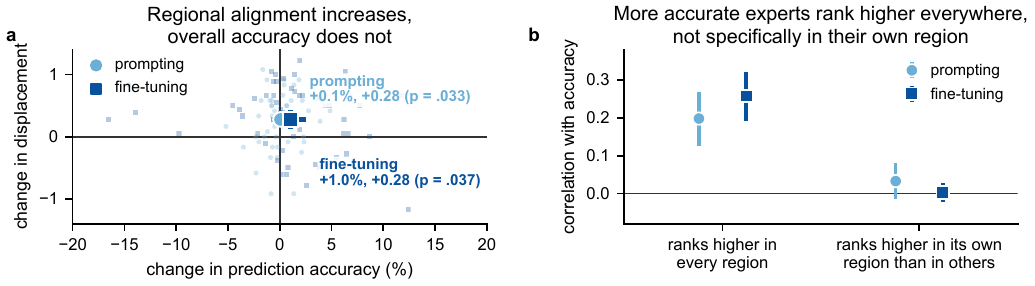}
\caption{\textbf{Specialization improves regional alignment without changing overall alignment.} \emph{(a)} Turning LLaMA into experts: change in prediction accuracy against displacement, one point per LPP subject; the base model, read as six identical copies, has displacement 0. \emph{(b)} Correlation across a family's six experts between accuracy and (left) rank over all six regions or (right) how much higher an expert ranks in its own region than in the others; mean and 95\% interval over 62 subjects. Only the right-hand quantity enters displacement.}
\label{fig:dissociation}
\end{figure}

Specializing a model changes where it aligns without changing how well it predicts the brain overall (\Cref{fig:dissociation}). On the 49-subject dataset, turning LLaMA into a family of experts leaves its whole-brain prediction accuracy nearly unchanged (+0.1\% under prompting and +1.0\% under fine-tuning; $p$ = .88 and .18) but gives the family a displacement of +0.28 under both interventions ($p$ = .033 and .037, read at a fixed set of layers; \Cref{fig:dissociation}a). Overall accuracy does not explain which expert aligns with which region either. Within a family, a more accurate expert ranks somewhat higher (r = +0.20 and +0.26 across the 62 subjects), but in every region alike, and such a shared advantage cancels out of displacement. Whether an expert ranks higher in its own region than in the others, which is what displacement counts, is unrelated to its overall accuracy (r = +0.03 and +0.00, $p$ = .18 and .82; \Cref{fig:dissociation}b). This may explain why instruction tuning does not appear to improve brain alignment at matched model size when evaluated across the whole brain \citep{gao2025increasing}: specialization changes which region each expert predicts best, which a whole-brain average does not register.

\section{Conclusion}\label{sec:conclusion}

We show that specializing a language model toward a cognitive domain, whether by a system prompt or by fine-tuning, moves its representations toward the brain system associated with that domain, across base models and datasets, while control interventions without cognitive content do not. This holds for domains whose brain systems lie largely outside the language network, and it comes without any gain in overall prediction accuracy. The fact that a prompt, which changes no weights, works about as well as fine-tuning suggests that specialization amplifies a representational structure that is already present in a model trained only on text. Specialized models thus offer a way to probe which parts of broader human cognition a language model may have obtained; our work provides a methodology to test whether these emergent representations align with functionally specialized subsystems in the human brain.

\begin{center}\rule{0.5\linewidth}{0.5pt}\end{center}

\section*{AI use statement}

We used LLM-based assistants to implement and debug analysis code, to give feedback on the experimental design and controls, and to draft and edit text and figures. As part of the method, Meta-Llama-3-8B-Instruct generated the distilled fine-tuning commentary (\Cref{app:experts}), and Qwen2.5-32B-Instruct and Mistral-Small-24B-Instruct served as blind judges of the prompted experts' generations (\Cref{app:behavior}). We did not use generative AI to collect, preprocess or alter any brain data. Every number was regenerated from scripts and checked by the authors. We take responsibility for the final content of this work, including text, claims or artifacts produced with the aid of generative AI.

\section*{Ethics statement}

This study uses three previously published fMRI datasets \citep{lebel2023natural,li2022petit,pereira2018toward} and collects no new human data. Each was collected with informed consent under the approval of its own institution's review board and is publicly released in de-identified form. All models we prompt or fine-tune are publicly available. The compute the study used is reported in \Cref{app:environment}.

\section*{Reproducibility statement}

All three fMRI datasets are public, as are the three base models and every tool in the pipeline. The six system prompts and the composition of the fine-tuning data are given in \Cref{app:experts}; the region definitions, the encoding pipeline and both outcome statistics in \Cref{app:brain}; and the statistical tests, including the treatment of hyperparameters, in \Cref{app:inference}. Every per-configuration displacement test is an exact permutation test over the 720 relabelings of six experts to six regions, or 120 over five networks, so no displacement $p$-value depends on a sampling seed. We will release the code as supplementary material; it implements every step of the method, including the prompts, the construction of the fine-tuning data and the training configurations, the behavioral validation, the encoding models, the region definitions and the alignment statistics.

\bibliography{iclr2027_conference}
\bibliographystyle{iclr2027_conference}

\appendix
\crefalias{section}{appendix}

\section{The Six Cognitive Domains}\label{app:cognitive-domains}

The six domains are chosen to cover the cognitive terms that have reliable meta-analytic maps in Neurosynth \citep{yarkoni2011large}, with each domain bringing together several near-synonymous terms that converge on one construct. Sensory processing covers the primary sensory modalities; spatial processing covers spatial relations, navigation and visual scenes; numerical processing covers calculation and arithmetic; reasoning covers inference, planning and cognitive control; social processing covers social interaction and reasoning about other minds; and the abstract domain covers abstract concepts, meaning and values. The granularity is set by where the meta-analytic evidence supports distinct and reliable cortical localization: fewer domains would merge dissociable systems, such as spatial navigation and early sensory processing, and more would subdivide below the resolution these maps support. The same six domains define the experts' prompts, their fine-tuning data and the brain regions they are tested against, so each expert has exactly one region it is expected to align with (\Cref{fig:overview}).

\section{Additional Details About Constructing and Validating the Expert Models}\label{app:experts}

\paragraph{Domain prompts.} Each prompted expert receives one of the following system prompts, held fixed across every stimulus and every dataset. Base models whose chat template carries no system role receive the same text prepended to the user turn.

\begin{itemize}
\tightlist
\item \emph{Sensory:} ``You are an expert in sensory and physical processing. Focus on concrete sensory experiences, physical actions, bodily sensations, textures, temperatures, sounds, tastes, and smells.''
\item \emph{Spatial:} ``You are an expert in spatial and visual processing. Focus on spatial relationships, visual scenes, locations, navigation, distances, shapes, colors, and movement through space.''
\item \emph{Numerical:} ``You are an expert in numerical and temporal reasoning. Focus on quantities, numbers, time sequences, durations, ordering of events, mathematical relationships, and logical sequences.''
\item \emph{Reasoning:} ``You are an expert in reasoning and planning. Focus on causal relationships, logical inference, problem solving, goal-directed behavior, strategy, and decision making.''
\item \emph{Social:} ``You are an expert in social cognition and communication. Focus on social interactions, emotions, intentions, beliefs, desires, interpersonal relationships, and communication patterns.''
\item \emph{Abstract:} ``You are an expert in abstract concepts and values. Focus on abstract ideas, moral judgments, ethical principles, cultural values, philosophical concepts, and symbolic meaning.''
\end{itemize}

\paragraph{Fine-tuning data.} Each domain's set combines up to three sources (\Cref{tab:ftdata}; sensory has no long-form text), all posed as prompt--completion pairs with the loss on the completion only. The first is public instruction-following and reading-comprehension corpora, filtered to the domain by source, with items whose answers collapse onto a template removed. Sensory is carried by the physical-goal items of PIQA \citep{bisk2020piqa} together with creative-writing prompts; social by SocialIQa \citep{sap2019socialiqa}, persona-grounded dialogue \citep{zhang2018personalizing} and the dialogues of ProsocialDialog \citep{kim2022prosocialdialog}, with a smaller mix of counseling and open-domain conversation; numerical by the word problems of GSM8K \citep{cobbe2021training} and Orca-Math \citep{mitra2024orca} and the time-sensitive questions of TimeQA \citep{chen2021dataset} and MultiTQ \citep{chen2023multi}; spatial by NarrativeQA \citep{kovcisky2018narrativeqa} and passages of fiction; reasoning by the open-domain instruction data of OpenHermes-2.5; and abstract by reflective prompts on quotations and essays.\footnote{Corpora without an accompanying publication are identified by their Hugging Face names:
\url{https://huggingface.co/datasets/ChaoticNeutrals/Creative_Writing-ShareGPT},
\url{https://huggingface.co/datasets/AlekseyKorshuk/fiction-books},
\url{https://huggingface.co/datasets/teknium/OpenHermes-2.5},
\url{https://huggingface.co/datasets/Abirate/english_quotes},
\url{https://huggingface.co/datasets/sgoel9/paul_graham_essays},
\url{https://huggingface.co/datasets/manu/project_gutenberg},
\url{https://huggingface.co/datasets/euirim/goodwiki},
\url{https://huggingface.co/datasets/open-r1/Mixture-of-Thoughts}, and
\url{https://huggingface.co/datasets/wikimedia/wikipedia} (snapshot 20231101.en).} For sensory, social and numerical these tasks are themselves about the domain; for spatial, reasoning and abstract the material is more general, and the other two sources add text written about the domain. The second source is long-form text that contains the domain's vocabulary, posed as continuation: the first half of a passage, after a one-line focus such as ``Continue this passage focusing on time, dates, durations, numbers, and measurements,'' is the input and the second half the target. It draws on PG-19 \citep{rae2019compressive}, Project Gutenberg and GoodWiki for abstract; English Wikipedia for spatial and numerical; GovReport \citep{huang2021efficient} for numerical; Mixture-of-Thoughts and LongWriter \citep{bai2025longwriter} for reasoning; and SODA \citep{kim2023soda} and HH-RLHF \citep{bai2022training} for social. The third source is distilled domain commentary. The same passages from NarrativeQA, fiction and essays were given, for every domain, to Meta-Llama-3-8B-Instruct \citep{grattafiori2024llama} with a system prompt describing that domain's content (for the numerical domain, for example, explicit times, durations, counts and measurements) and asking for an 80- to 200-word commentary rich in it, inventing plausible domain details where the passage offers few. The training pair takes the passage alone as the prompt and the commentary as the completion, so in this source the domain is carried by the target text rather than by an instruction.

\begin{table}[t]
\caption{\textbf{Composition of each domain's fine-tuning set.} Rows after deduplication. Instruction corpora are the filtered public sets; distilled are the teacher-generated domain commentaries.}
\label{tab:ftdata}
\centering
\small
\setlength{\tabcolsep}{4pt}
\begin{tabular}{lrr>{\raggedright\arraybackslash}p{4.4cm}rr}
\toprule
Domain & Instruction & Distilled & Long-form additions & Train & Validation \\
\midrule
Sensory    & 17,843 & 3,673 & none & 21,516 & 557 \\
Spatial    & 14,916 & 3,668 & Wikipedia geography 2,850 & 21,434 & 647 \\
Numerical  & 35,741 & 3,670 & Wikipedia biographies 2,852; GovReport 1,898 & 44,161 & 1,172 \\
Reasoning  & 12,147 & 3,674 & Mixture-of-Thoughts 2,849; LongWriter 1,336 & 20,006 & 660 \\
Social     & 47,039 & 3,674 & SODA 2,844; HH-RLHF helpful 1,906 & 55,463 & 1,402 \\
Abstract   &  2,811 & 3,667 & PG-19 5,711; Gutenberg 1,765; GoodWiki philosophy 1,607 & 15,561 & 821 \\
\bottomrule
\end{tabular}
\end{table}

\paragraph{Adapter training.} All three base models are the instruction-following releases (Meta-Llama-3-8B-Instruct, Qwen3-8B and Phi-4). LoRA adapters are applied to all attention and feed-forward projection matrices at rank 8, 16 or 64 ($\alpha$ twice the rank) and trained for at most three epochs. Training on domain text can buy domain fit at the cost of general language ability, which would confound the brain analyses, since a model that has simply become worse would predict brain responses differently for reasons unrelated to its domain. Checkpoints are therefore selected on held-out domain loss, with a small penalty for any rise in perplexity on held-out general text, among checkpoints whose general-text perplexity stays below 1.30 times the base model's; every adapter used here stayed at or below 1.07 times. The remaining optimization settings are part of the released code.

\section{Additional Details About Measuring Brain Alignment}\label{app:brain}

\paragraph{Brain regions.}
For any term used in published studies, Neurosynth \citep{yarkoni2011large} gives a whole-brain z-map of how much more reliably each voxel is reported active in studies that mention the term than in studies that do not. We therefore defined each domain's region from a set of domain-specific terms:
\begin{itemize}
  \item \textbf{spatial}: spatial, visuospatial, navigation, place;
  \item \textbf{numerical}: calculation, arithmetic;
  \item \textbf{reasoning}: reasoning, control, cognitive control, executive,
        planning, rule, solving;
  \item \textbf{social}: social, social cognition, social interaction, mind,
        theory of mind, face;
  \item \textbf{abstract}: abstract, concept, conceptual, semantic, semantic
        memory;
  \item \textbf{sensory}: auditory cortex, visual cortex, somatosensory cortex,
        olfactory, early visual.
\end{itemize}

For the five non-sensory categories, the constituent term maps were combined by taking, at each voxel, the maximum z-value across all terms in that category. This voxel-wise maximum operation was chosen (rather than, e.g., averaging) so that a voxel implicated strongly by any one term would be represented in the category map, without weaker or noisier terms diluting that signal. The resulting category-level combined map was then thresholded to retain a fixed-size region of interest, defined as a fixed set of the top 5,000 voxels by z-value.

The sensory domain needs one extra step, because its terms differ widely in statistical strength and the strongest would otherwise crowd out the rest. We first threshold each sensory term's map to its own top 1,200 voxels, which gives every modality the same number of candidate voxels, and then combine these masks by voxelwise maximum as for the other domains. Because the masks overlap, the result can exceed 5,000 voxels (it has 5,641), so we keep its top 5,000 by z-value. The sensory region thus matches the others in size while still drawing on every modality. The regions overlap little (Dice 0.00 to 0.04 for most pairs; at most 0.23, between numerical and reasoning). Each region is defined on the common grid of Le Petit Prince and projected into each LeBel subject's functional space with the dataset's own transforms (1,432 to 1,825 cortical voxels per region).

The result does not depend on region size. We re-ran every LeBel and Le Petit Prince configuration with two alternative sizes per dataset: 45--70 and 200--306 voxels per region on LeBel, and the top 10\% and the full support of each map on Le Petit Prince. All 24 readings give a positive displacement.

\begin{figure}[t]
\centering
\includegraphics[width=0.9\linewidth]{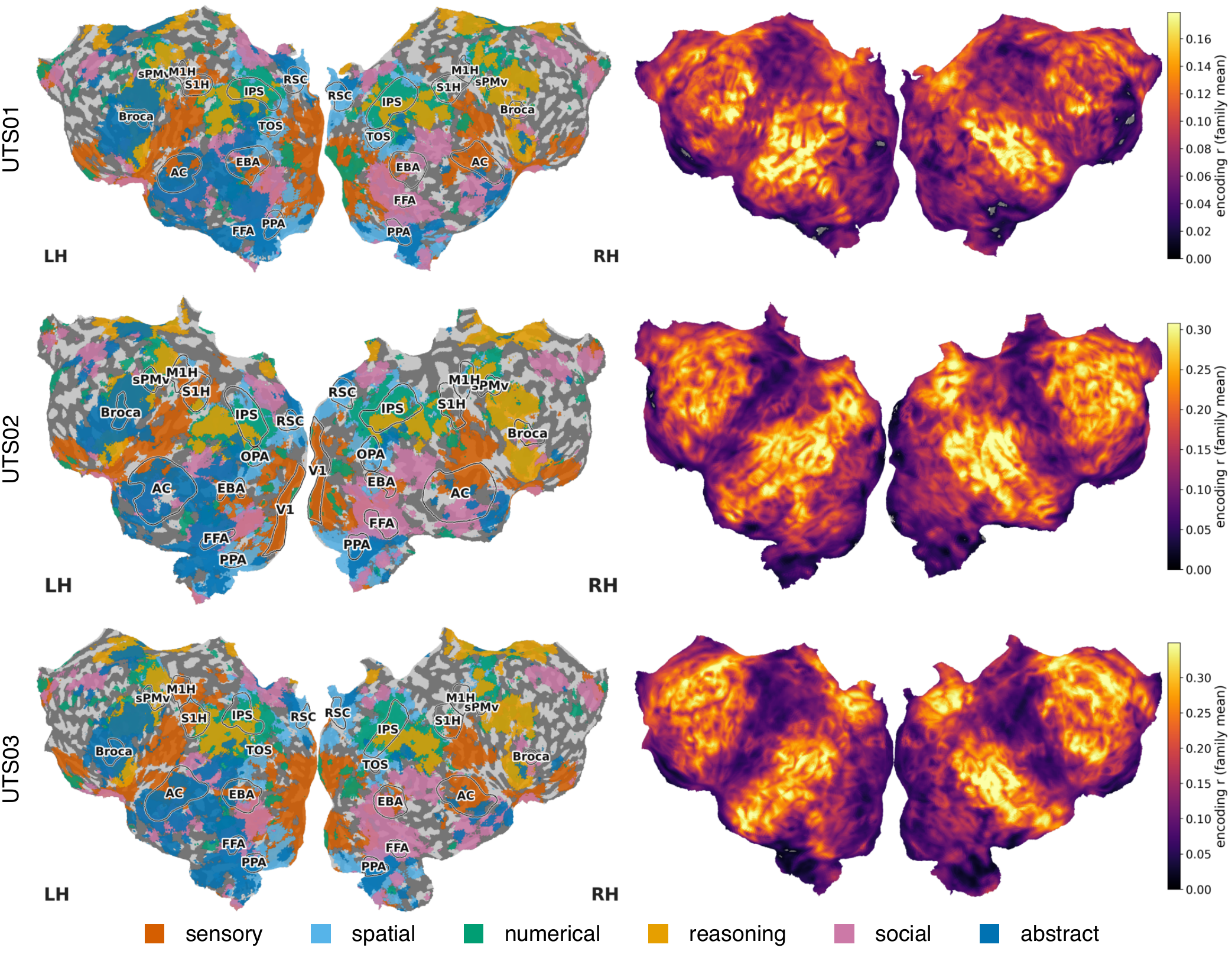}
\caption{\textbf{The six regions in every LeBel subject.} Left: the Neurosynth regions for the six domains on each subject's flattened cortical surface, with the dataset's localizer and atlas regions outlined. Right: the prediction accuracy of the prompted LLaMA family (mean over its six experts), each subject on its own color scale. Right (RH) and left (LH) hemispheres are shown.}
\label{fig:regions}
\end{figure}

\paragraph{Functional networks on Pereira.} The Pereira release's five functional networks replace the Neurosynth regions,
each mapped to the domain closest to its best-established function: language \citep{fedorenko2011functional} to abstract
meaning, multiple-demand \citep{fedorenko2013broad} to reasoning, default-mode to social cognition, visual to spatial processing, and auditory to sensory processing \citep{power2011functional}. The last two mappings are the loosest for a reading task. Since no network corresponds to numerical processing, we drop the numerical expert on this dataset, so displacement is 3.0 minus the matched expert's mean rank among the five, tested against an exact null over all 120 relabelings.

\paragraph{Participants.} The three LeBel subjects are UTS01 to UTS03 of the public release \citep{lebel2023natural}, the Le Petit Prince cohort is the English-language one (n = 49; \citealp{li2022petit}), and the ten Pereira subjects \citep{pereira2018toward} are those who completed the passage-reading experiments (experiments 2 and 3).

\paragraph{Encoding models.} Each context is represented by the hidden state at the final token of the templated input, the position from which the model would begin its reply. Prompted experts receive their instruction as a system message; fine-tuned experts receive none, and read the text as a plain user turn on the narrative datasets and as raw text on Pereira. On LeBel, each subject's 93 to 98 training stories and three test stories follow the standard split, and the feature PCA and the ridge penalty are fitted on the training stories. The penalty is selected per response component by bootstrap cross-validation over 20 log-spaced values from $10^{1}$ to $10^{8}$. Le Petit Prince, with shorter runs and a 2 s TR, uses a lighter variant: each word is represented by its 25-word context read at the end of the templated turn, features are binned to TRs, reduced to 512 principal components and given four delays, and a ridge model with a fixed penalty of $10^{5}$ is fit and scored with leave-one-run-out cross-validation. Pereira presents isolated sentences rather than a continuous scan, so its features are sentence-level with no delay stage, cross-validation is five-fold over passages so that sentences of one passage never fall on both sides of a split, and the ridge grid spans $10^{2}$ to $10^{6}$. One LeBel test story is presented ten times, which gives a noise ceiling per voxel, the square root of the Spearman--Brown corrected split-half reliability of the repeats; it is reported only as context (\Cref{app:alignment}). The results do not depend on the response-PC stage. Keeping more response variance (about 50\% or 70\% instead of 100 components) leaves displacement unchanged (r = 0.93 and 0.89 with 100 components under the same encoder), and fitting the ridge regression to every voxel directly, without response PCs, lowers prediction accuracy by about 7\% but leaves all six LeBel configurations positive, with displacements that track the reported ones. Further, the alignment does not depend on the moments when the stimulus is about a region's domain: scoring each region only on the held-out timepoints whose content matches its domain, labeled with the question-answering features of \citet{singh2025evaluating}, does not raise displacement in the running example. The experts' advantage thus comes from how each represents the language as a whole, not only from the parts of the stimulus that concern its domain.

\paragraph{Outcome statistics.} Let $r_e(v)$ be the encoding accuracy of expert $e$ at voxel $v$ (the correlation between predicted and observed responses on held-out data), and $d_e(v) = r_e(v) - \frac{1}{K}\sum_{e'} r_{e'}(v)$ its difference from the family mean, with $K = 6$ experts ($K = 5$ on Pereira). For region $R_i$ and the rest of the cortical mask $\overline{R_i}$, the contrast
\[
c_{e,i} \;=\; \frac{1}{|R_i|}\sum_{v \in R_i} d_e(v) \;-\; \frac{1}{|\overline{R_i}|}\sum_{v \notin R_i} d_e(v)
\]
is how much better than its siblings expert $e$ predicts inside region $i$ than outside it. Let $\rho_{e,i} \in \{1,\dots,K\}$ be expert $e$'s rank on this contrast within region $i$ (1 = largest), and let $\pi(i)$ be the expert paired with region $i$, by theory the expert for that region's domain.

\emph{Displacement} is
\[
D(\pi) \;=\; \frac{K+1}{2} \;-\; \frac{1}{K}\sum_{i=1}^{K} \rho_{\pi(i),\,i},
\]
so that 0 is chance ($(K+1)/2$ is the mean of the ranks $1,\dots,K$: 3.5, or 3.0 on Pereira) and the maximum, $(K-1)/2$, is reached when every region's own expert ranks first. Its $p$ is the fraction of all $K!$ pairings $\pi$ whose displacement is at least that of the theory-given pairing. The out-of-region term makes displacement specific by construction: an expert that predicts all of cortex better than its siblings gains nothing, since the same advantage is subtracted on both sides.

\emph{$\Delta r$} keeps the magnitude that ranking discards and uses only the voxels inside each region. Let $m_{e,i} = \frac{1}{|R_i|}\sum_{v \in R_i} r_e(v)$ be expert $e$'s mean accuracy inside region $i$; then $m_{\pi(i),i}$ is that of the region's own expert, and the other $K-1$ experts are averaged:
\[
\Delta r \;=\; \frac{1}{K}\sum_{i=1}^{K}\Big(m_{\pi(i),\,i} \;-\; \frac{1}{K-1}\sum_{e \neq \pi(i)} m_{e,i}\Big),
\]
one value per subject, tested across subjects with a one-tailed t-test. $\Delta r$ is reported for its size and its test across subjects. It is small in absolute terms because the six experts are one model specialized six ways, whose prediction maps correlate at r $\approx$ 0.98 in the median configuration.

Both statistics are computed in each subject at each layer; when a configuration is read over a band of layers they are averaged over its layers, since ranking is not linear and averaging maps first would give a different answer, and then over subjects.

\paragraph{Control families.} The control families of \Cref{sec:experts} are \textbf{random-LoRA} adapters, untrained and matched in norm to the cognitive adapters; \textbf{seed-only} fine-tunes, six adapters trained by the identical recipe on equally sized samples of one corpus pooled from all six domains, each with its own seed, so that they differ only in sampling and initialization; \textbf{random-prompt} families, the six prompt slots filled with expert roles in unrelated fields (for example zebra striping anatomy or vintage typewriter restoration; all prompts are listed with the released code); \textbf{surface-form} fine-tunes on English Wikipedia sentences rewritten by six fixed rules that change form but not content (leetspeak, alternating case, reversed words, disemvoweling, uppercase emphasis and pirate dialect); and the \textbf{keyword} family, bare lists of each domain's most distinctive vocabulary with no instruction. A control family has no correct assignment of members to regions, so its expected displacement is 0 and it is scored under every assignment, at the same setting as the expert family it is compared with.

\section{Statistical Inference}\label{app:inference}

\paragraph{Tests.} Every per-configuration displacement $p$ is the exact relabeling test: the subjects' mean displacement is computed under each of the 720 assignments of the six experts to the six regions (120 on Pereira; the same assignment for every subject), and $p$ is the fraction scoring at least as high, so the smallest attainable $p$ is 1/720 (1/120 on Pereira). It needs no sampling and makes no assumption about the subjects, who are all scored against the same six model fits. $\Delta r$, which is continuous, is tested across subjects with a one-tailed t-test. The pooled rows of \Cref{tab:main} standardize each subject's displacement by the spread of its own relabeling null, so that six-expert and five-expert datasets share one scale, and test the pooled mean against a Monte Carlo null that samples one relabeling per dataset and configuration.

\paragraph{Hyperparameters.} Two choices in the pipeline are hyperparameters: the layer at which representations are read out (layer 6, 12, 18, 24 or 30, or the mean over a shallow band of layers 6 and 12 or a deep band of layers 18 to 30) and, for fine-tuning, the adapter's rank (8, 16 or 64). As with any hyperparameter, we swept them, and each configuration in \Cref{tab:main} is reported at its best setting. The default rank is 64. Everything else, including the regions and both statistics, is fixed for all configurations.

Choosing the best setting can by itself raise a result, so we checked that the table holds up once this choice is taken into account. Put through the same sweep and rule, experts relabeled at random reach a mean displacement of +0.31 and the intervention-matched control families +0.23, against +0.55 for the actual experts ($p$ = .0002 and .002). Without any selection, reading every configuration at every layer, 66 of the 90 readings are positive ($p$ = .004).

\paragraph{The eighteen configurations.} Each base model, intervention and dataset is a replication in its own right, and all eighteen are reported whatever their outcome. The evidence is the pattern across them, which the test above establishes while relabeling configurations on the same dataset together, since they share subjects. The individual $p$-values in \Cref{tab:main} are those of the reported setting and show where the effect is clearest; ten of the twelve below .05 remain so after a false-discovery-rate correction across the eighteen (Benjamini--Hochberg, $q$ = .05).

\section{Additional Details on the Behavioral Validation}\label{app:behavior}

The two interventions express specialization differently, so each is read by the measure closest to it: perplexity on held-out domain text for fine-tuning, the objective the adapters were trained on, and the content of what the experts generate for prompting, which leaves the weights untouched and acts only while the instruction is present.  

\begin{figure}[t]
\centering
\includegraphics[width=\linewidth]{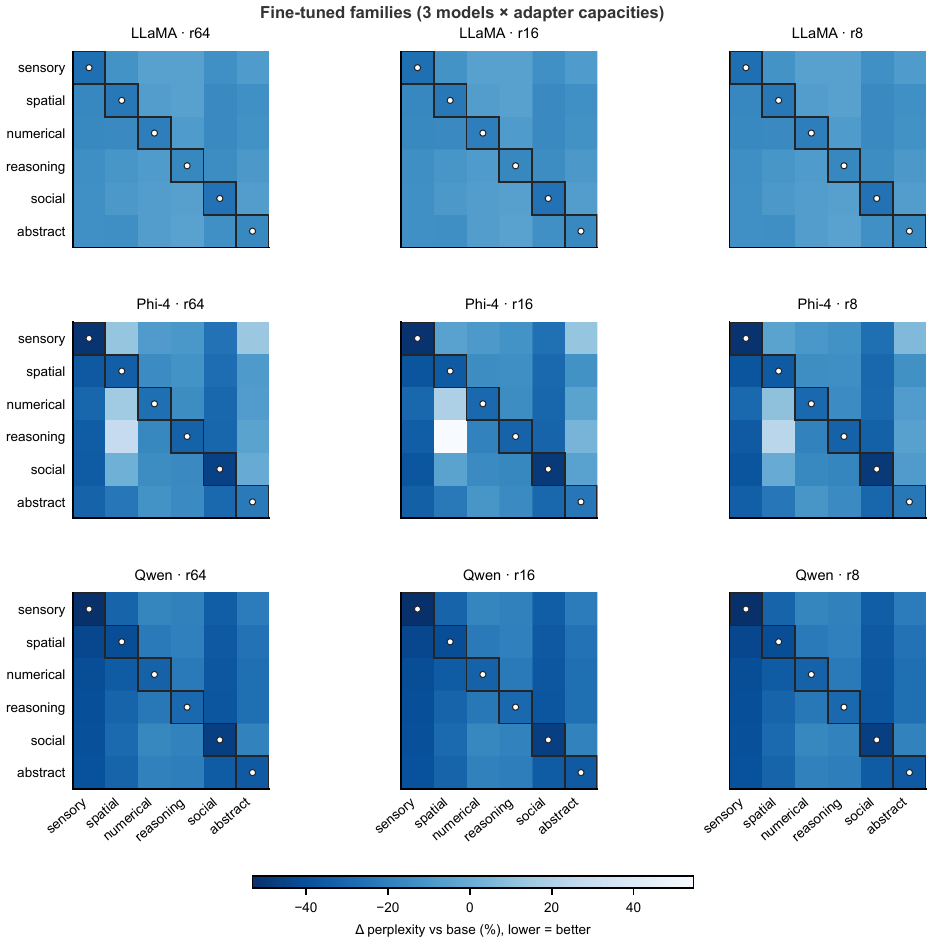}
\caption{\textbf{The perplexity matrix of every fine-tuned family.} Percentage change in perplexity relative to the base model, each expert (rows) on each domain's held-out text (columns), for the three base models at three adapter ranks; the white dot marks each column's strongest improvement.}
\label{fig:ppl_all}
\end{figure}

\paragraph{Perplexity matrices.} \Cref{fig:ppl_all} gives the perplexity matrix of every fine-tuned family, all three base models at all three adapter ranks. The matrix is read down its columns: holding one domain's text fixed and comparing the six experts on it poses the same six-way question as the brain analysis and avoids comparing domains whose text differs in how predictable it is. In each of these nine base model $\times$ rank combinations, each domain's held-out text is improved most by its own expert, with mean own-domain improvements of 21.6\% to 39.6\%. How much more each expert improves its own domain's text than the other experts do orders the base models Phi-4, LLaMA, Qwen (21.5, 13.5 and 10.3 percentage points), the same order as their mean fine-tuning displacement over the three datasets (+0.54, +0.51 and +0.37).

\begin{figure}[t]
\centering
\includegraphics[width=\linewidth]{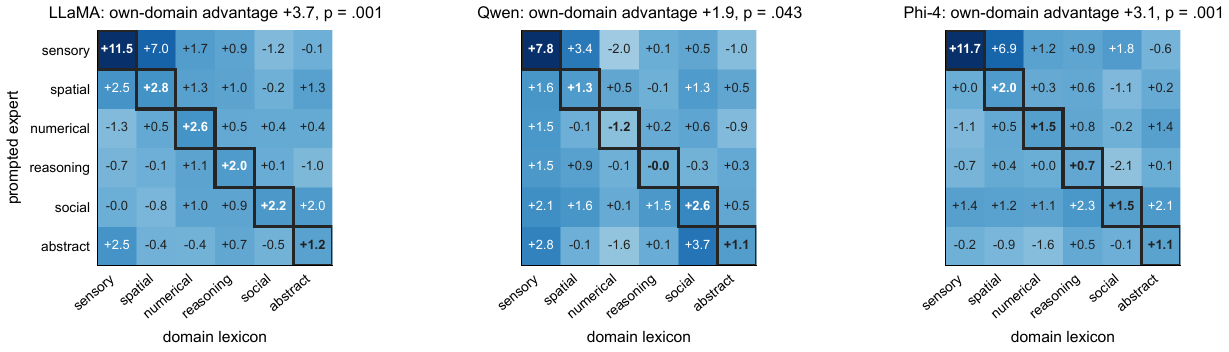}
\caption{\textbf{Lexicon share of generated text for every base model.} As \Cref{fig:behavior}b, for LLaMA, Qwen and Phi-4.}
\label{fig:lexicon_all}
\end{figure}

\paragraph{Lexicon shift.} \Cref{fig:lexicon_all} gives the lexicon-shift matrix of \Cref{fig:behavior}b for all three base models. Each expert continues the same 24 story openings from the LeBel stimuli. Each domain's lexicon is the 150 content words most specific to that domain's fine-tuning validation text (by TF-IDF), after removing stopwords and every word that appears in any of the six system prompts. A cell is the share of an expert's generated words that belong to a domain's lexicon minus the share for the base model, in words per 100. The statistic is the mean of the six matched cells, tested against the 720 relabelings of prompts to lexicons: 3.7, 1.9 and 3.1 more own-domain words per 100 for LLaMA, Qwen and Phi-4 ($p$ = .001, .043 and .001).

\begin{figure}[!htbp]
\centering
\includegraphics[width=\linewidth]{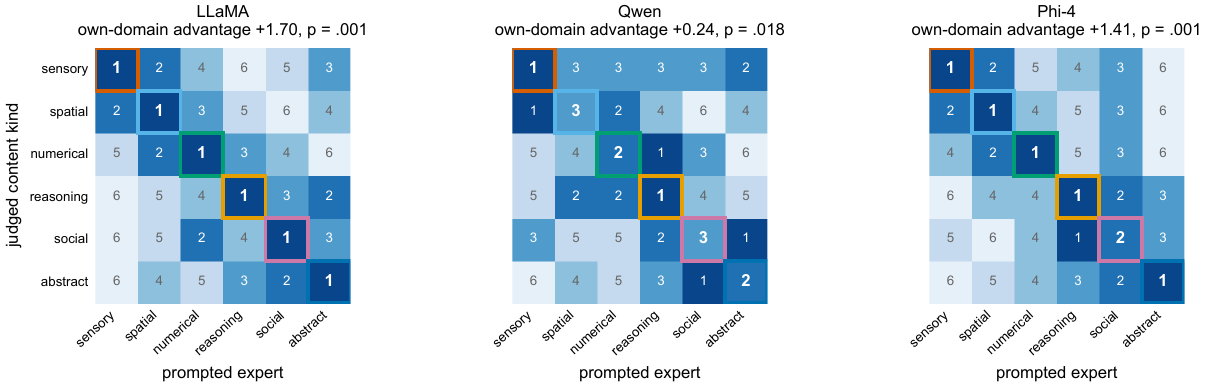}
\caption{\textbf{The blind judge's ranking for every base model.} Each cell is the expert's rank on that content kind (1 = the judge scores this expert's continuations highest of the six, relative to the base model); the outlined diagonal is the matched cell.}
\label{fig:judge_all}
\end{figure}

\paragraph{The blind judge.} \Cref{fig:judge_all} gives the judge matrix of \Cref{fig:behavior}c for all three base models. The judge, Qwen2.5-32B-Instruct \citep{qwen2025qwen25technicalreport} at temperature 0, rates each continuation from 0 to 10 on how strongly it engages each of six kinds of content, each described in one line (for example, ``numerical and temporal content: numbers, quantities, counting, durations, dates, times and sequences''), and is told to ignore meta-commentary; it sees no model name, prompt or domain label. Sentences that announce the instruction are removed before judging; judging the full text instead leaves every conclusion unchanged (own-domain advantage +1.66, +0.66 and +1.55, all $p$ = .001). A matrix entry is the judged score minus the same judge's score for the base model's continuation of the same opening, and the own-domain advantage is the mean matched entry minus the mean of the other entries, tested with the same 720-relabeling permutation. The lexicon construction and the judge's full prompt are released with the code. A second judge from a different model family, Mistral-Small-24B-Instruct\footnote{\url{https://huggingface.co/mistralai/Mistral-Small-24B-Instruct-2501}}, replicates the result: LLaMA +1.96 with six of six content kinds and Phi-4 +1.63 with five of six, both $p$ = .001, with Qwen in the same direction (+0.14, $p$ = .088; +0.54 on the full text, $p$ = .028). Neither judge favors its own family: both rank the three base models in the same order, and the Qwen-family judge scores the Qwen experts lowest.

\paragraph{Benchmark battery.} The battery groups MMLU \citep{hendrycks2020measuring} subject tests into the six domains, fixed before any model was run: conceptual physics, anatomy and nutrition for sensory; high-school geography and astronomy for spatial; elementary and high-school mathematics and high-school statistics for numerical; formal logic and logical fallacies for reasoning; sociology, professional psychology and human sexuality for social; and moral scenarios, moral disputes and philosophy for abstract. On this battery, the expert prompted toward a domain is ahead of the other five on that domain's questions in LLaMA and Qwen (battery displacement +1.17 and +0.67) and at chance in Phi-4 (0.00), +0.61 on average, a directional result ($p$ = .11) beside the two significant content-based measures.

\clearpage
\section{Additional Details on the Regional Alignment Result}\label{app:alignment}

\begin{figure}[t]
\centering
\includegraphics[width=\linewidth]{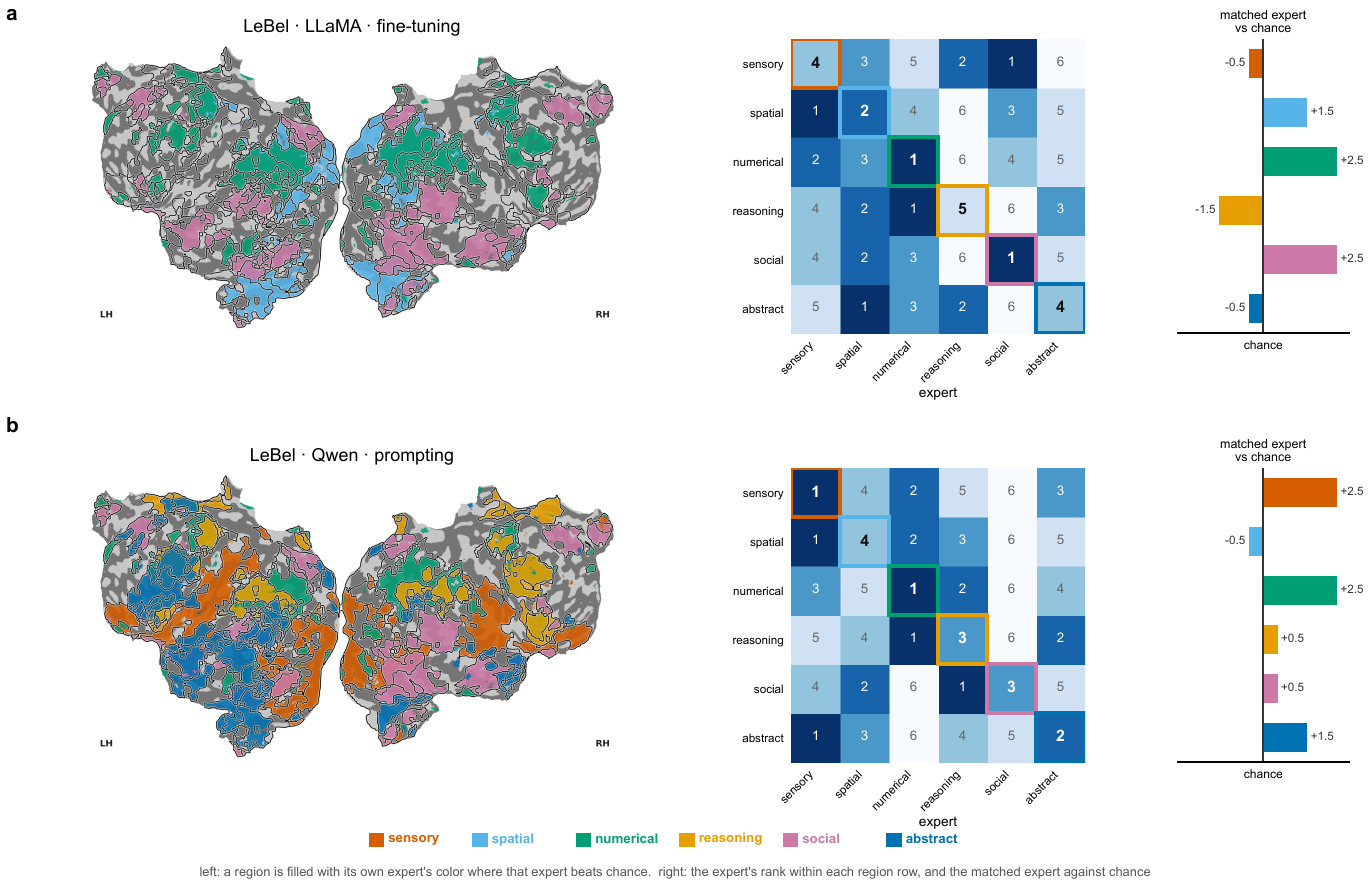}
\caption{\textbf{LeBel, LLaMA fine-tuning and Qwen prompting.} Each row is one configuration in the format of \Cref{fig:lead}. Left: every region is outlined and labeled, and filled with its own expert's color when that expert beats chance inside it, with the expert's rank printed beside the region name. Right: the expert $\times$ region matrix, where each cell prints that expert's rank within the region's row and the region's own expert is outlined, and beside it the matched expert's rank against chance.}
\label{fig:cells_lebel_llama}
\end{figure}

\begin{figure}[t]
\centering
\includegraphics[width=\linewidth]{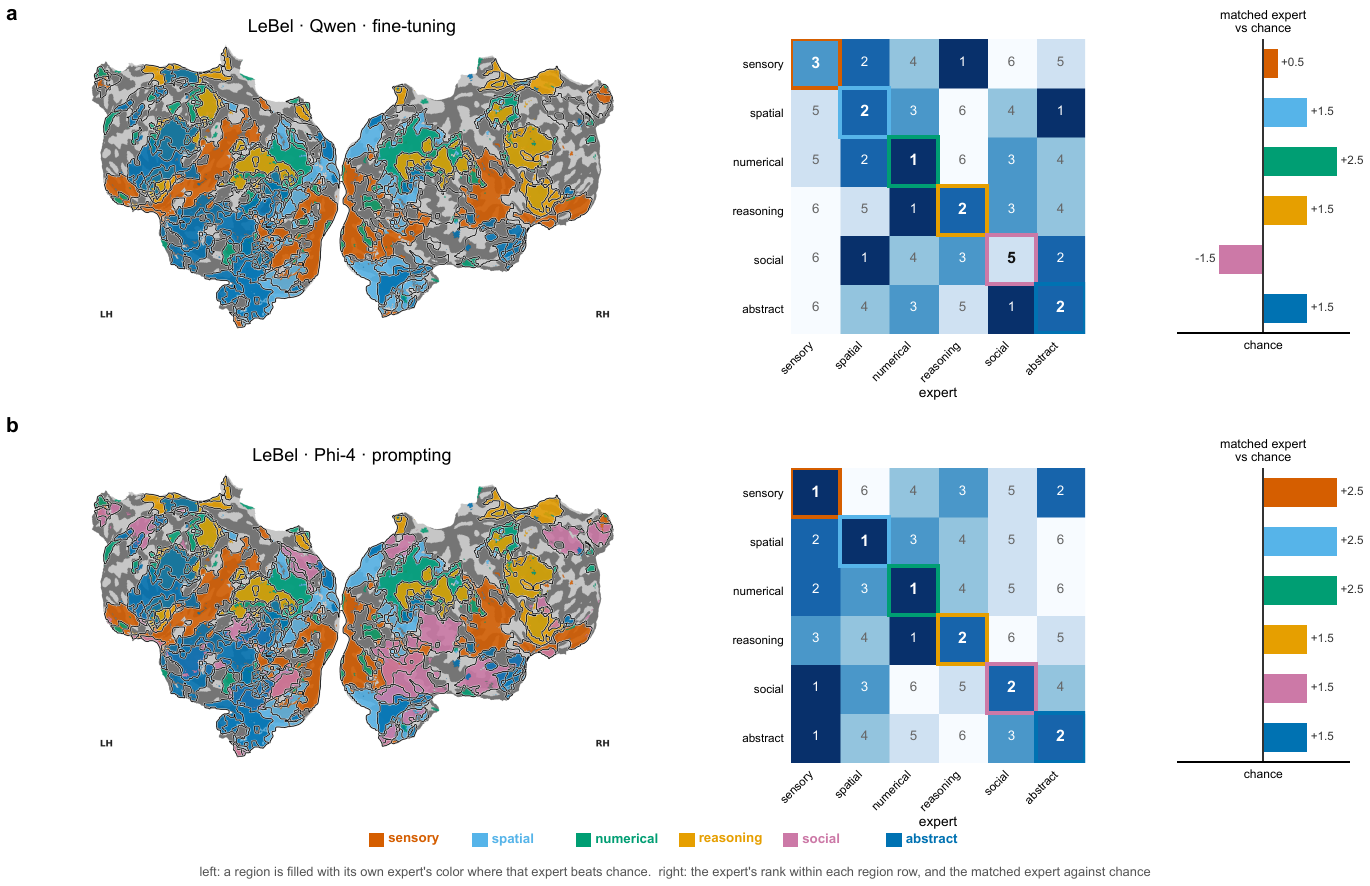}
\caption{\textbf{LeBel, Qwen fine-tuning and Phi-4 prompting.} Conventions as in \Cref{fig:cells_lebel_llama}.}
\label{fig:cells_lebel_qwen}
\end{figure}

\begin{figure}[t]
\centering
\includegraphics[width=\linewidth]{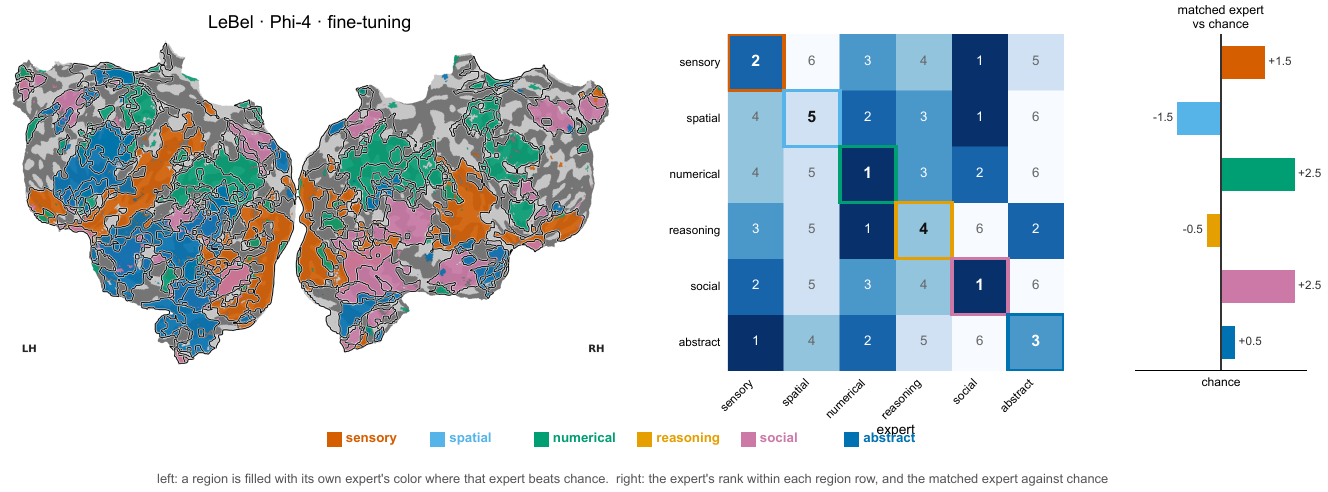}
\caption{\textbf{LeBel, Phi-4 fine-tuning.} Conventions as in \Cref{fig:cells_lebel_llama}.}
\label{fig:cells_lebel_phi4}
\end{figure}

\begin{figure}[t]
\centering
\includegraphics[width=\linewidth]{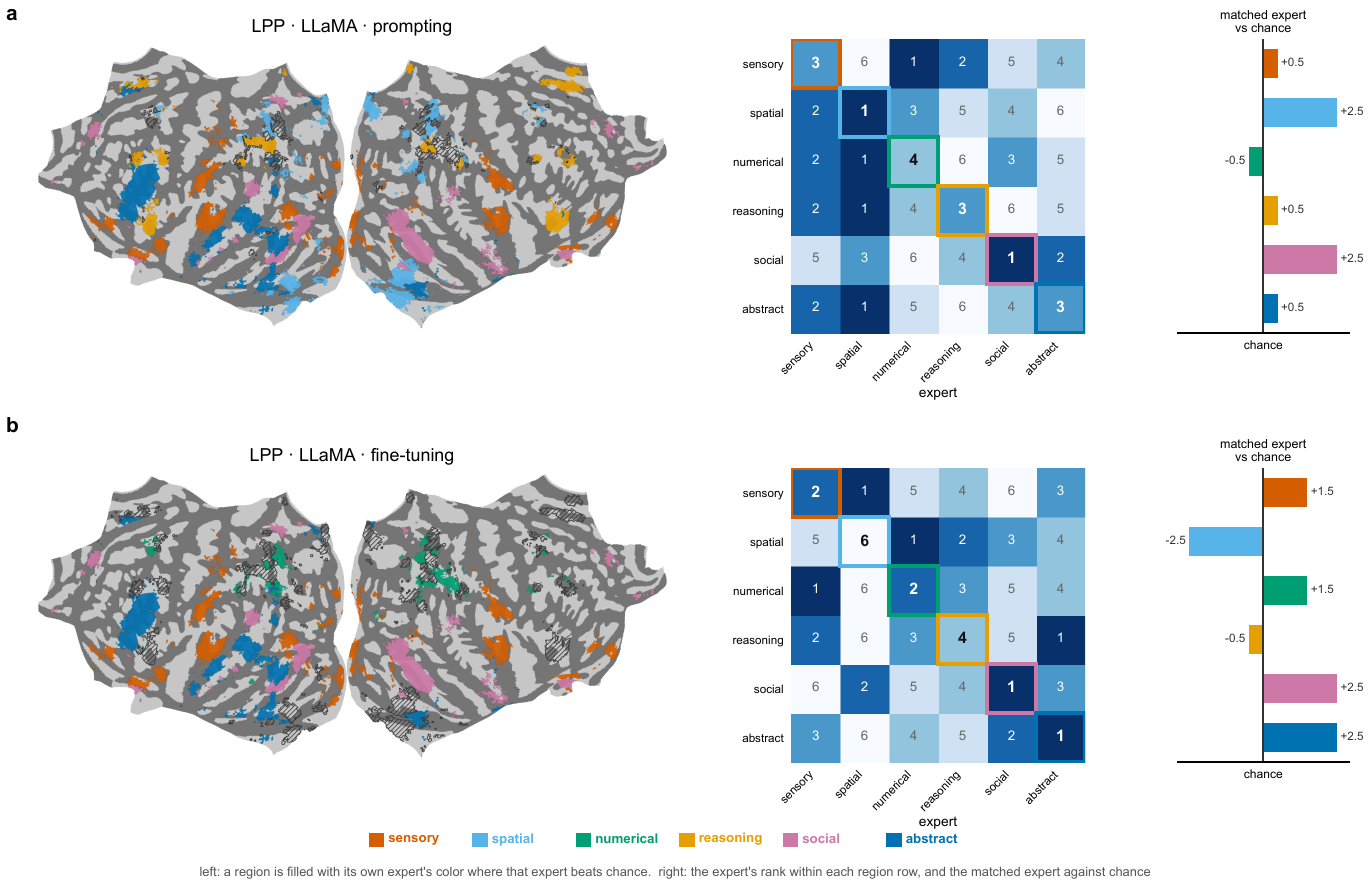}
\caption{\textbf{Le Petit Prince, LLaMA prompting and fine-tuning.} Conventions as in \Cref{fig:cells_lebel_llama}.}
\label{fig:cells_lpp_llama}
\end{figure}

\begin{figure}[t]
\centering
\includegraphics[width=\linewidth]{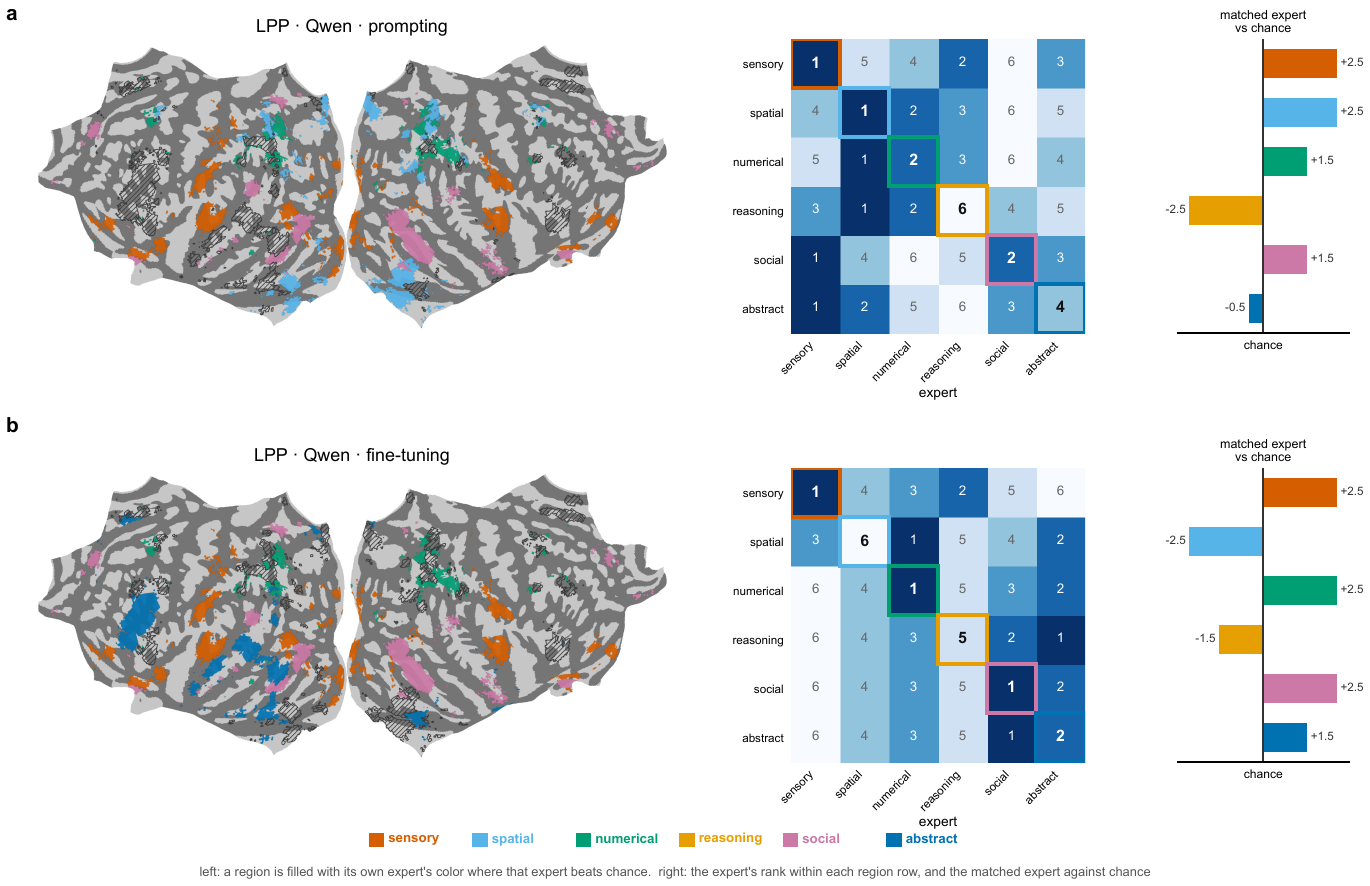}
\caption{\textbf{Le Petit Prince, Qwen prompting and fine-tuning.} Conventions as in \Cref{fig:cells_lebel_llama}.}
\label{fig:cells_lpp_qwen}
\end{figure}

\begin{figure}[t]
\centering
\includegraphics[width=\linewidth]{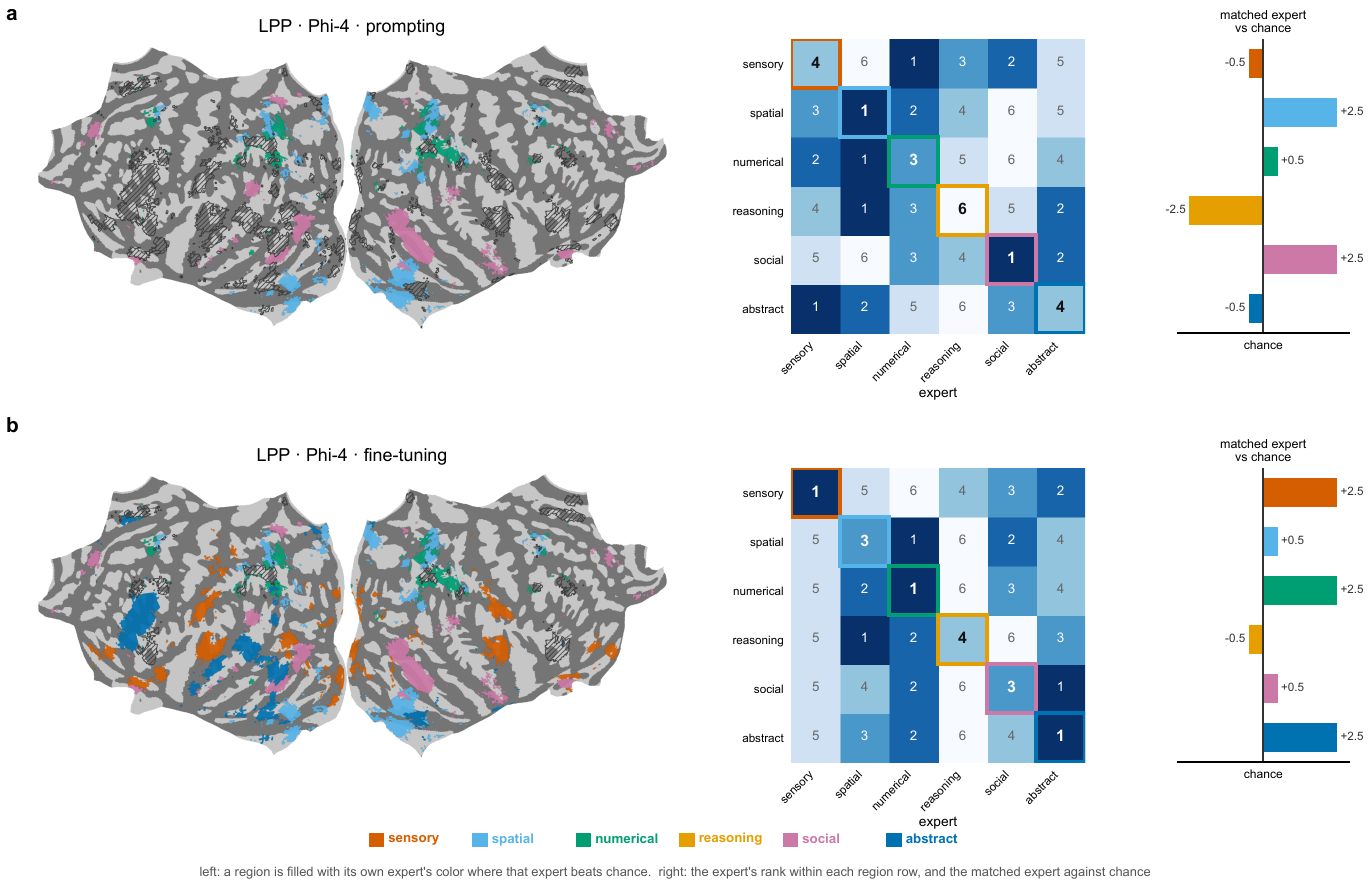}
\caption{\textbf{Le Petit Prince, Phi-4 prompting and fine-tuning.} Conventions as in \Cref{fig:cells_lebel_llama}.}
\label{fig:cells_lpp_phi4}
\end{figure}

\begin{figure}[t]
\centering
\includegraphics[width=\linewidth]{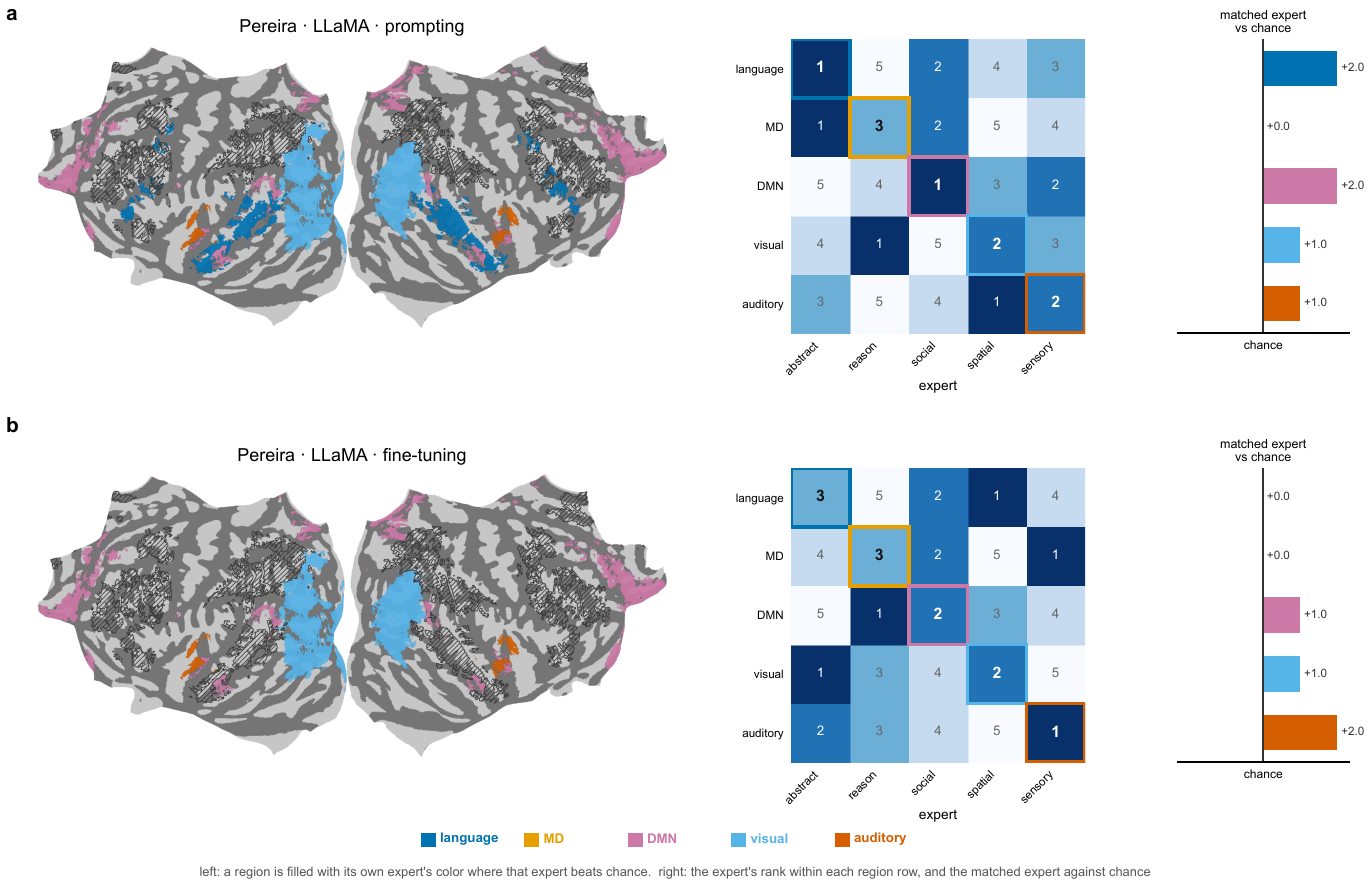}
\caption{\textbf{Pereira, LLaMA prompting and fine-tuning.} Conventions as in \Cref{fig:cells_lebel_llama}.}
\label{fig:cells_pereira_llama}
\end{figure}

\begin{figure}[t]
\centering
\includegraphics[width=\linewidth]{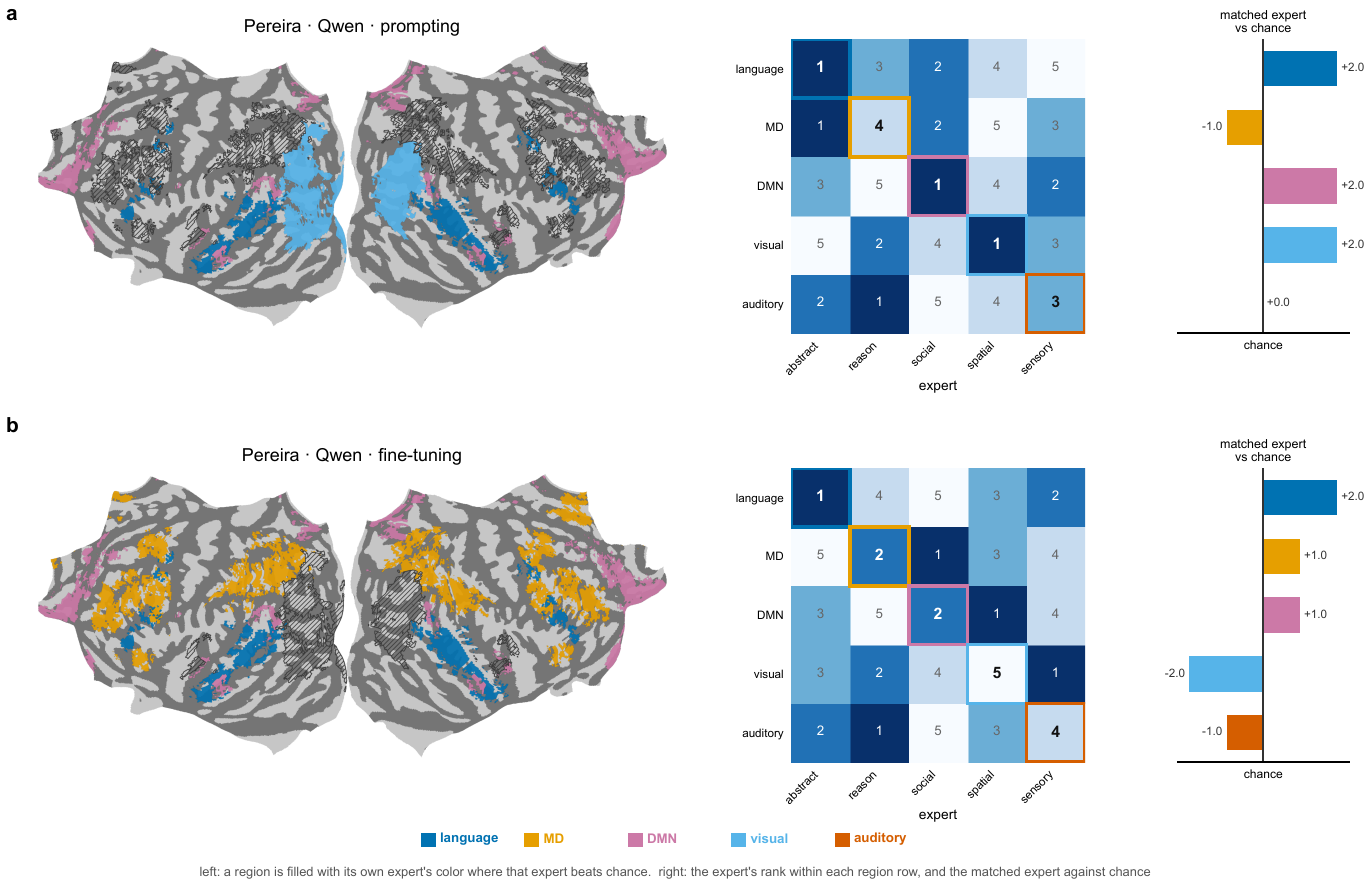}
\caption{\textbf{Pereira, Qwen prompting and fine-tuning.} Conventions as in \Cref{fig:cells_lebel_llama}.}
\label{fig:cells_pereira_qwen}
\end{figure}

\begin{figure}[t]
\centering
\includegraphics[width=\linewidth]{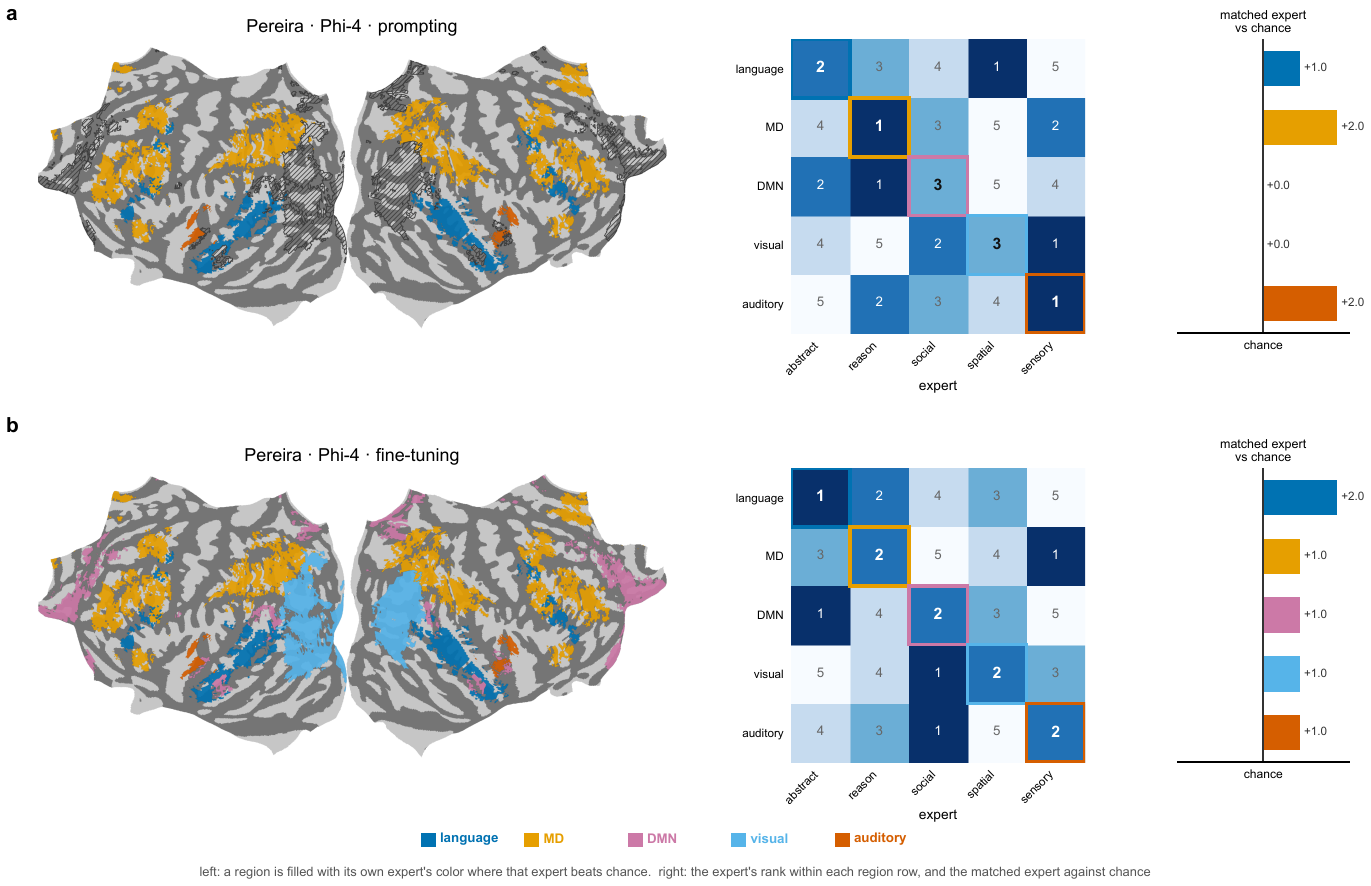}
\caption{\textbf{Pereira, Phi-4 prompting and fine-tuning.} Conventions as in \Cref{fig:cells_lebel_llama}.}
\label{fig:cells_pereira_phi4}
\end{figure}

\begin{figure}[!htbp]
\centering
\includegraphics[width=\linewidth]{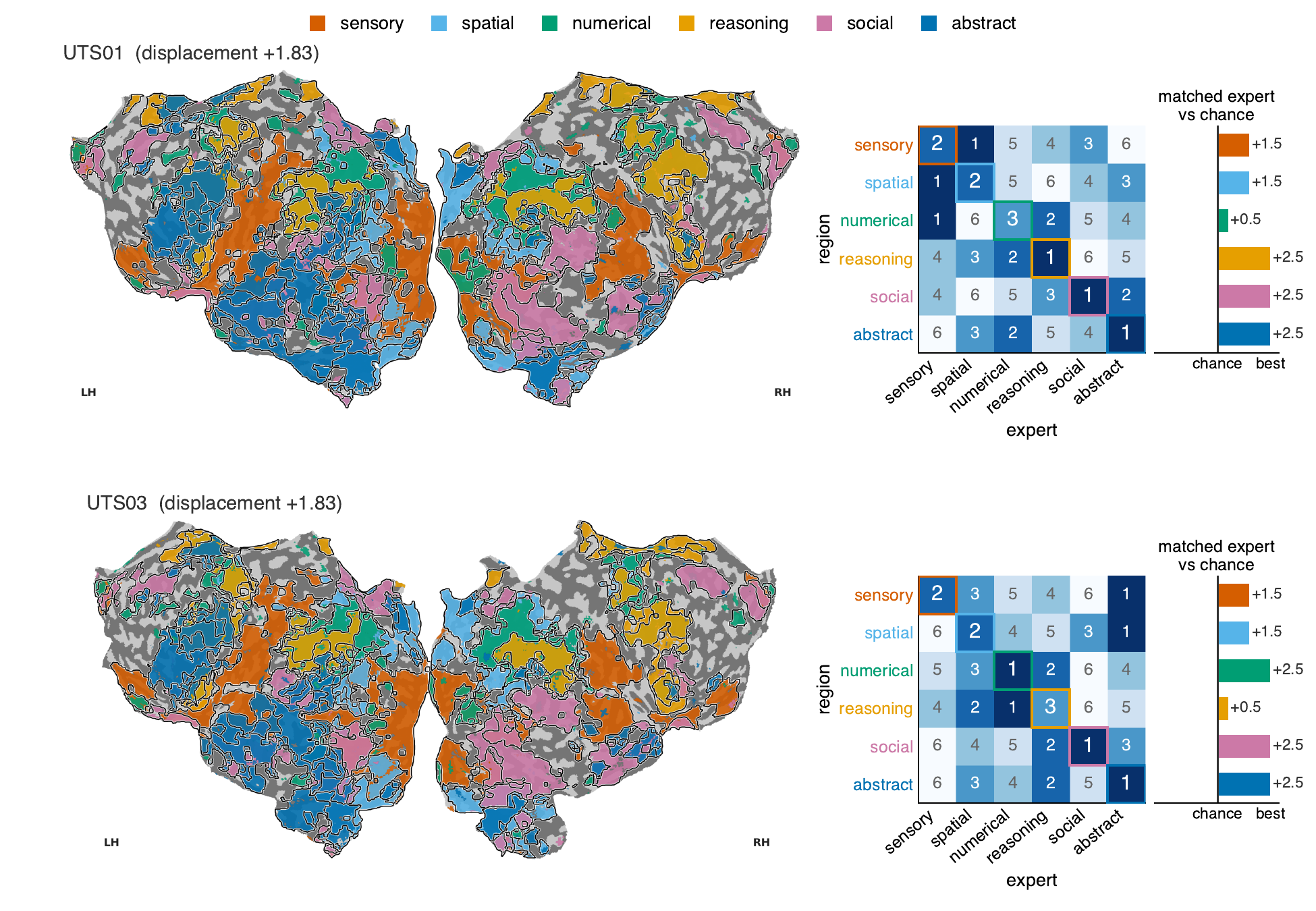}
\caption{\textbf{The running example in the other two LeBel subjects.} Each row as \Cref{fig:lead}, for UTS01 and UTS03, with each subject's own ranks: the region map, the region $\times$ expert matrix, and each matched expert's rank relative to chance.}
\label{fig:lead_subjects}
\end{figure}

\paragraph{Every configuration.} \Crefrange{fig:cells_lebel_llama}{fig:cells_pereira_phi4} show the other seventeen configurations of \Cref{tab:main} in the format of \Cref{fig:lead}: the cortical map on the left and the region $\times$ expert matrix on the right. Every configuration is positive and twelve are individually significant. The other six are the Qwen and Phi-4 prompting and Qwen fine-tuning configurations on Le Petit Prince and the three fine-tuning configurations on Pereira, where fine-tuning is the weaker intervention for every base model. LeBel's maps are drawn on its exemplar subject's surface; Le Petit Prince and Pereira define their regions once on a shared grid, so each of their configurations gives one group map. \Cref{fig:lead_subjects} shows the running example in the other two LeBel subjects.

\begin{figure}[t]
\centering
\includegraphics[width=0.9\linewidth]{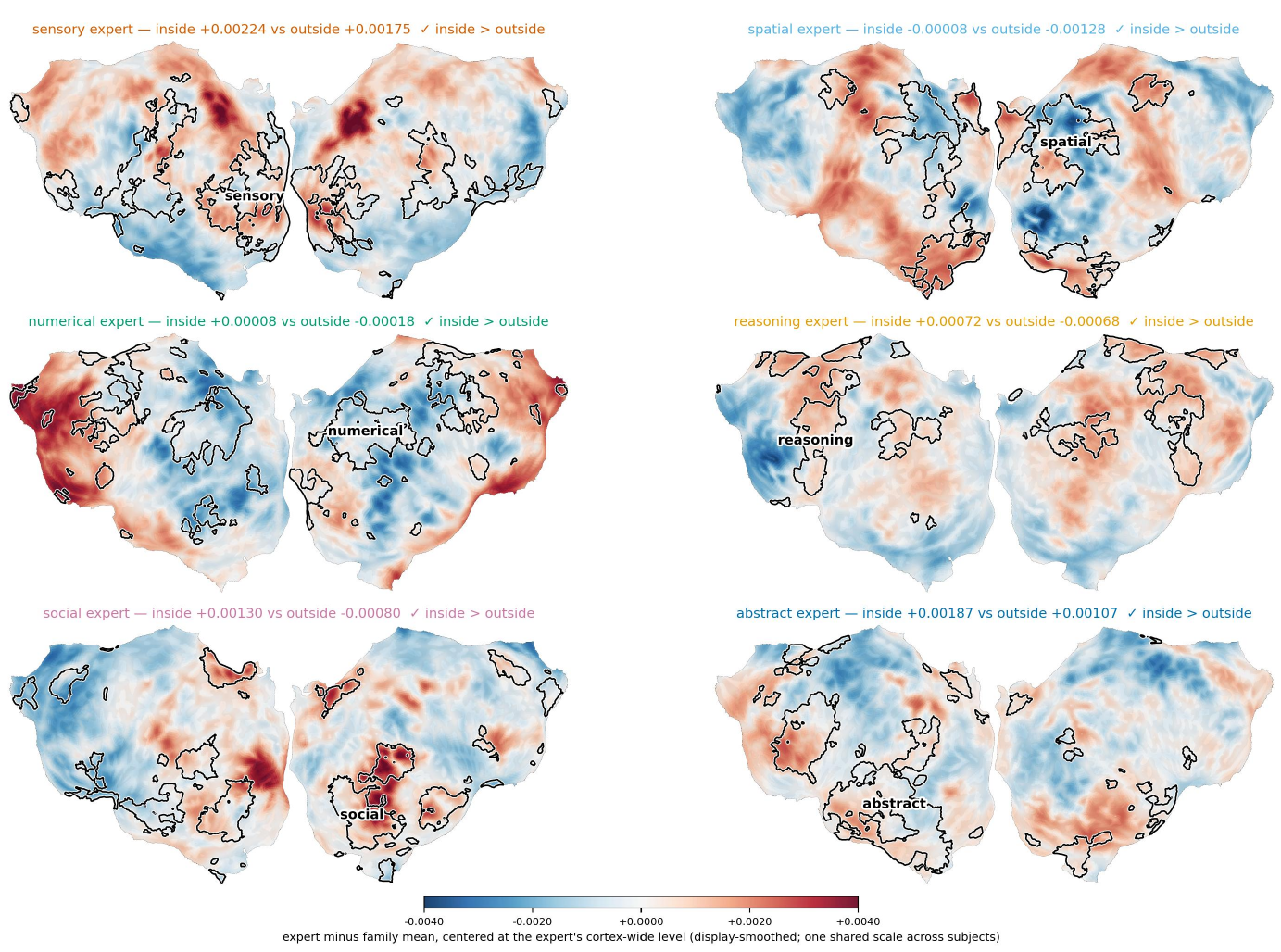}
\caption{\textbf{Per-expert differential maps (UTS01, LeBel, prompted LLaMA).} Each panel: one expert's encoding accuracy minus the family mean, on the cortical surface (smoothed for display), with that expert's own region outlined; for display, the color scale is centered at the expert's cortex-wide level. The printed values are the means of the unsmoothed map inside and outside the region, and the check mark records which is larger.}
\label{fig:expert_maps}
\end{figure}

\begin{figure}[t]
\centering
\includegraphics[width=0.87\linewidth]{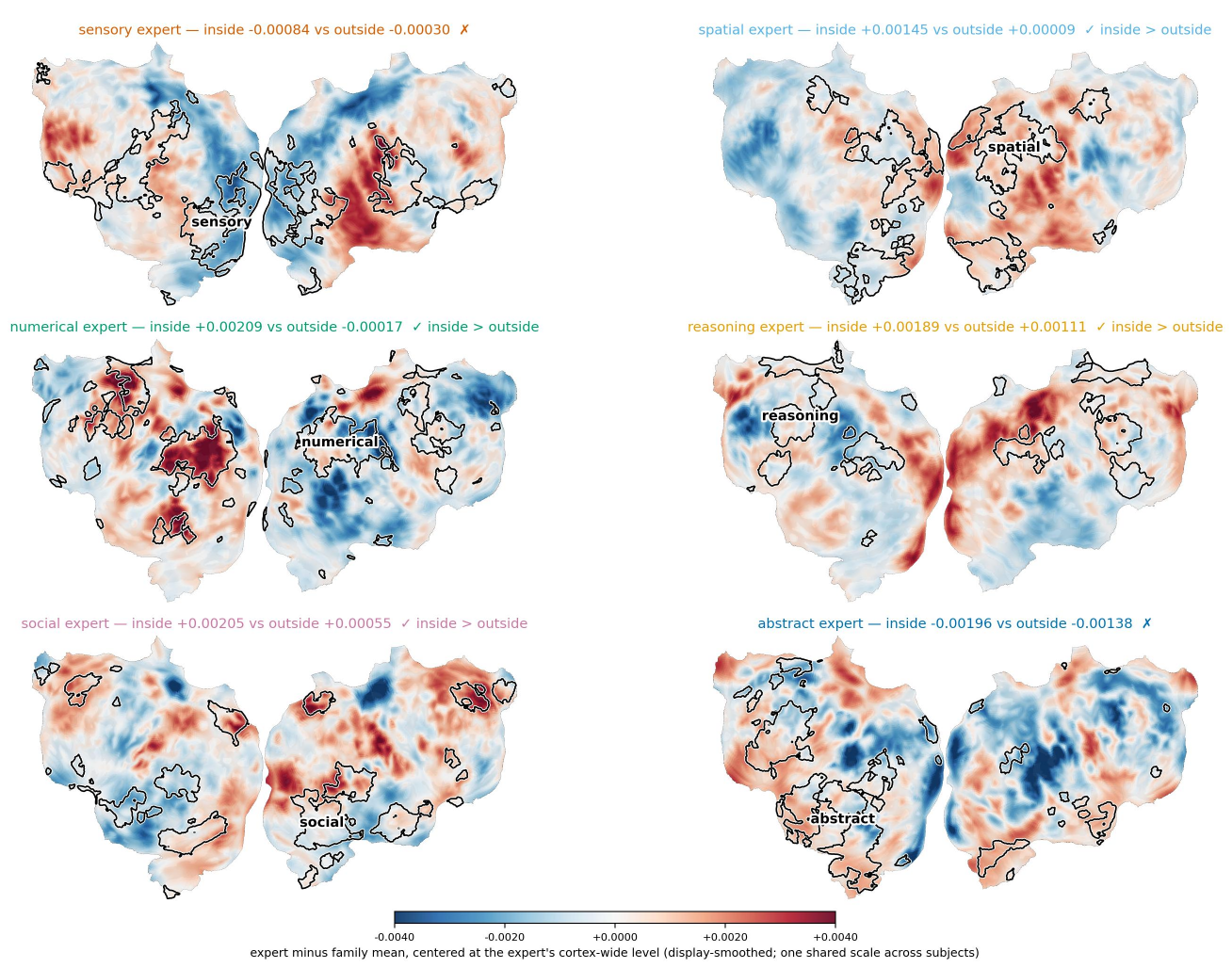}
\includegraphics[width=0.87\linewidth]{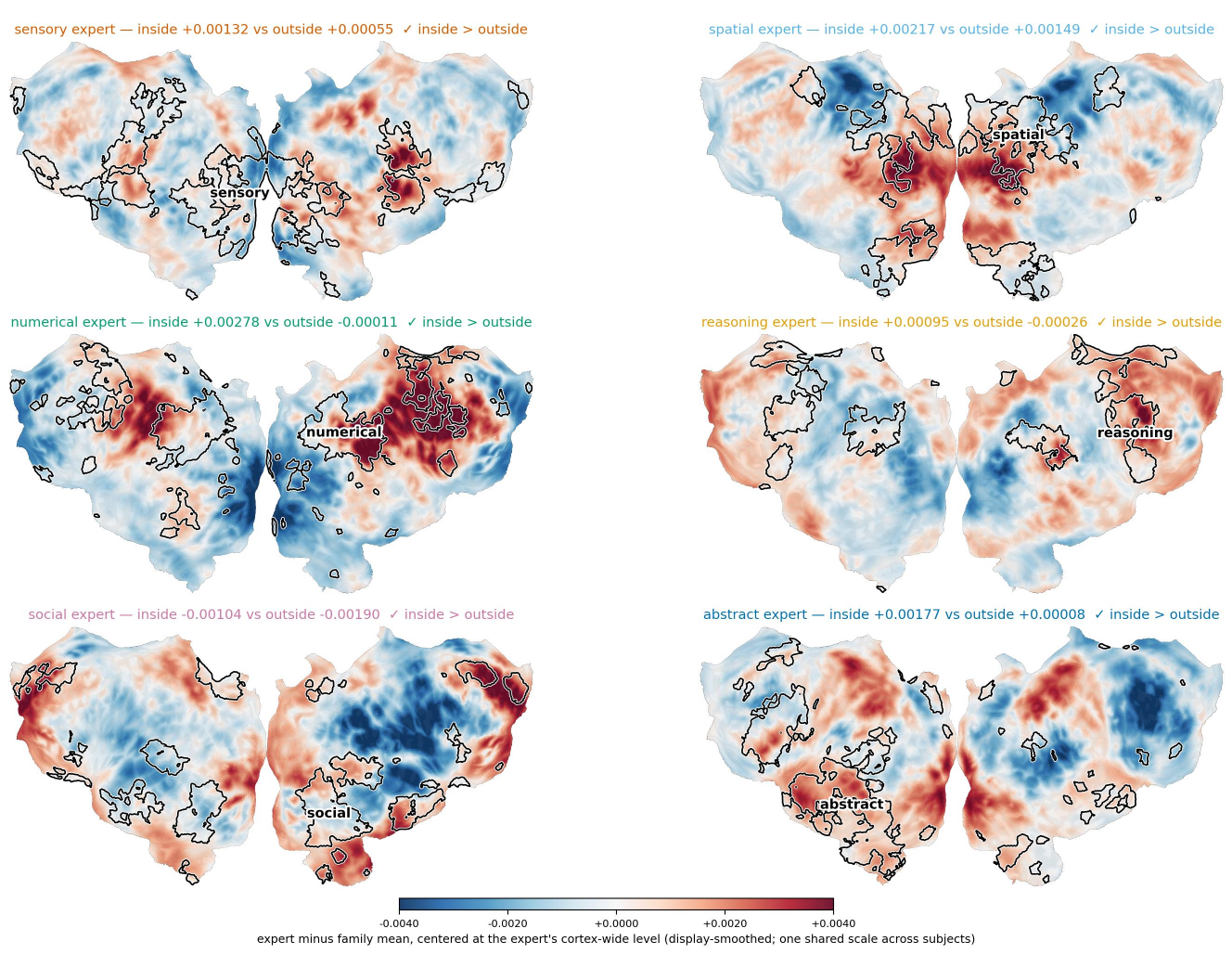}
\caption{\textbf{The same panels for UTS02 (top) and UTS03 (bottom).} Conventions as in \Cref{fig:expert_maps}.}
\label{fig:expert_maps_other}
\end{figure}

\paragraph{Per-expert cortical maps.} \Cref{fig:expert_maps} opens the running example on the cortex, one panel per expert: the expert's encoding map minus the family mean, with the expert's own region outlined. If the expert is aligned with its own region, the differential inside the outline should exceed the differential outside, which is the contrast displacement ranks. It does in all six panels of subject UTS01 and in sixteen of the eighteen expert $\times$ subject panels across the three subjects (\Cref{fig:expert_maps_other}; t(17) = 5.09, $p$ \textless{} .0001; sign test $p$ = .0007), and the verdicts hold when the outside term is restricted to the union of the other five regions (eighteen of eighteen) or when each hemisphere's mean is removed before scoring (all eighteen unchanged). In two panels both values are negative and the expert's deficit is smaller inside the region than outside.

\begin{figure}[t]
\centering
\includegraphics[width=0.6\linewidth]{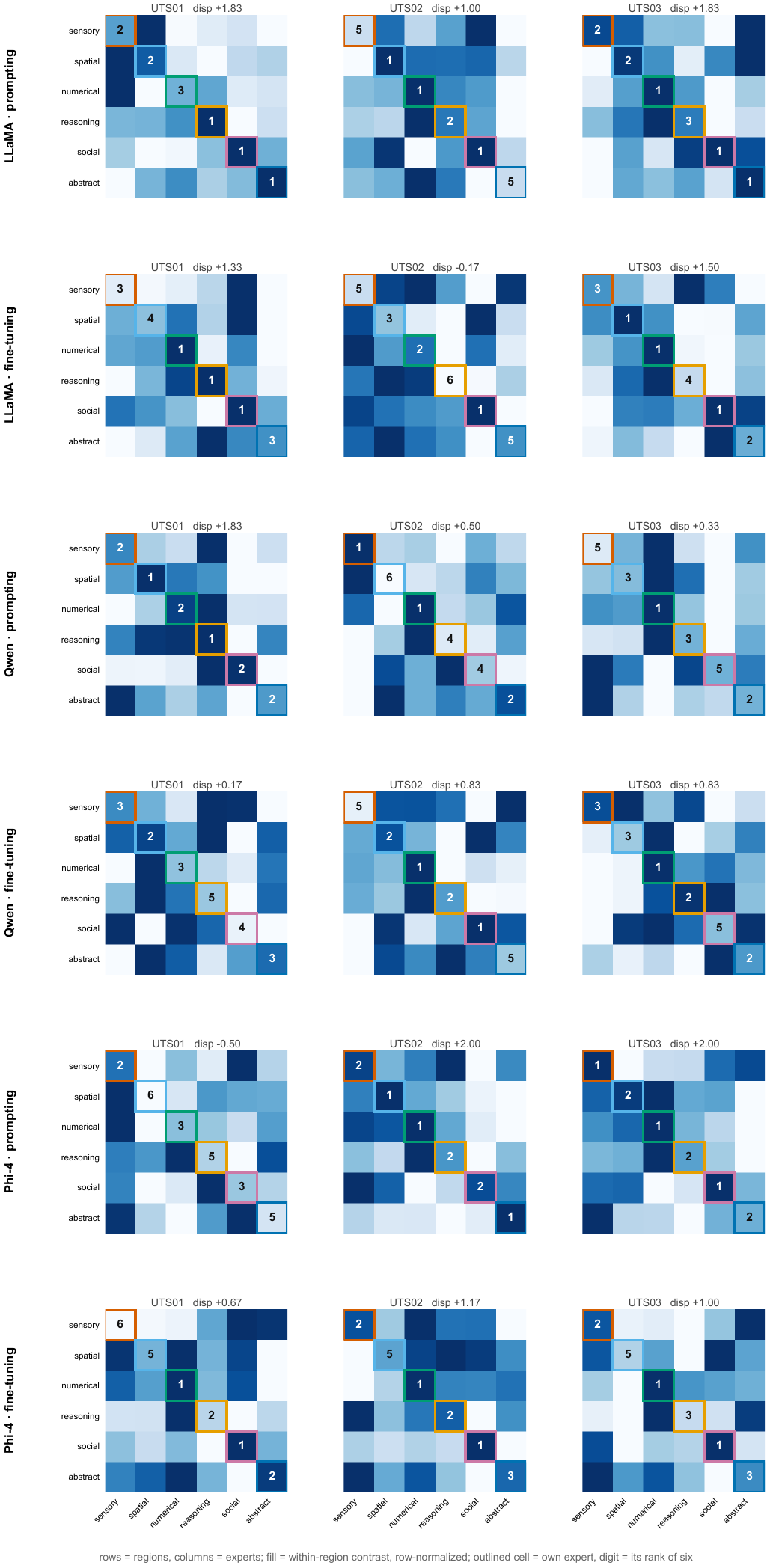}
\caption{\textbf{Per-subject region $\times$ expert matrices for the six LeBel configurations.} Each configuration at its \Cref{tab:main} setting; conventions as in \Cref{fig:lead}b, one panel per subject.}
\label{fig:subject_matrices}
\end{figure}

\paragraph{Per-subject matrices.} \Cref{fig:subject_matrices} gives the region $\times$ expert matrix for each LeBel subject separately, for all six LeBel configurations. Subject-level displacement is positive in sixteen of the eighteen family $\times$ subject cases, and in the two that are not, LLaMA fine-tuning in UTS02 and Phi-4 prompting in UTS01, the other two subjects are positive, so no group result is carried by one subject.

\begin{figure}[!htbp]
\centering
\includegraphics[width=\linewidth]{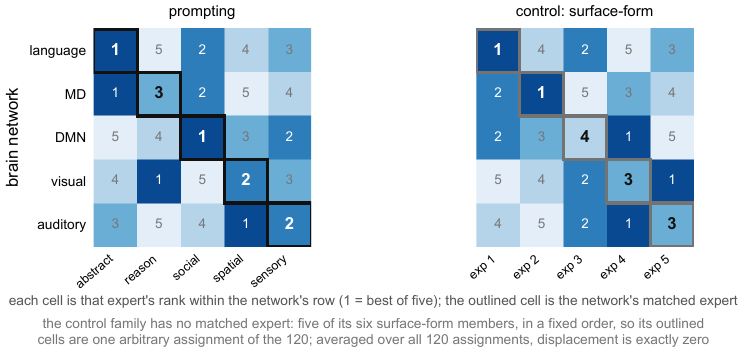}
\caption{\textbf{The network by expert matrices behind the Pereira numbers.} Left, LLaMA prompting; right, the surface-form control on the same base model. Rows are the five functional networks, columns the five experts with a matched network, and each cell prints that expert's rank within the network's row, with the network's own expert outlined. The surface-form control has no matched expert, so its outlined cells are one arbitrary assignment of its members, shown on identical conventions.}
\label{fig:pereira_matrices}
\end{figure}

\paragraph{Network by expert matrices on Pereira.} \Cref{fig:pereira_matrices} gives the matrices behind the Pereira rows of \Cref{tab:main}. Under LLaMA prompting the language network ranks the abstract expert first and the default-mode network ranks the social expert first, with the visual and auditory networks placing their matched experts second, and the surface-form control shows no such pattern.

\begin{figure}[!htbp]
\centering
\includegraphics[width=\linewidth]{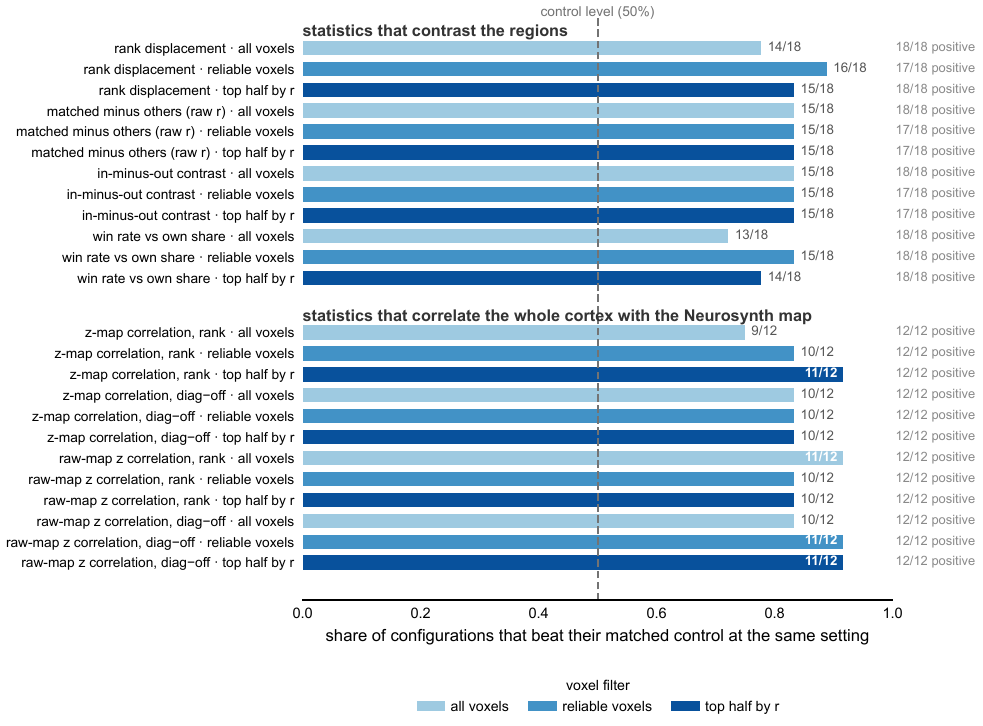}
\caption{\textbf{The result under every way of scoring it.} Eight scoring statistics $\times$ three voxel sets, for all eighteen configurations. Bars: how many cognitive configurations exceed their matched control (surface-form for fine-tuning, random-prompt for prompting, same dataset), each cognitive configuration at the layer and rank chosen under that statistic as for \Cref{tab:main} and its control, scored in its fixed member order, at the same setting; the dashed line is the even-odds level, against which the region statistics' worst result (13/18) has binomial $p$ = .048 and their best (16/18) $p$ = .0007. Right margin: every combination finds every computable configuration positive, with the one exception noted in the text, which sits at or just below zero. The whole-cortex statistics need continuous Neurosynth maps, which exist for LeBel and Le Petit Prince (twelve configurations); Pereira provides binary networks only.}
\label{fig:multiverse}
\end{figure}

\paragraph{Other ways of scoring.} \Cref{fig:multiverse} rescores every configuration with eight statistics, crossed with three ways of choosing voxels (every voxel, the voxels the family predicts above r = 0.05, and the better-predicted half of cortex). Four statistics are region-based: displacement; $\Delta r$; the \textbf{diagonal contrast}, each matched expert's inside-minus-outside score averaged over regions, so that a large lead in one region is not flattened by ranking; and the \textbf{win rate}, the share of a region's voxels best predicted by its own expert against that expert's share of the whole cortex. The other four ignore region boundaries and use the Neurosynth maps as continuous gradients, correlating each expert's accuracy map, voxel by voxel over the cortex, with each domain's z-map, and asking whether the matched correlations exceed the others (by rank and by difference, on differential and on raw accuracy maps). Every statistic finds every computable configuration positive on every voxel set, with one exception, Pereira Phi-4 fine-tuning, which sits at or just below zero under some statistics on the two filtered voxel sets, and the expert families beat their matched controls in thirteen to sixteen of the eighteen configurations under the region-based statistics and in nine to eleven of twelve under the whole-cortex ones.

\paragraph{Encoding performance against the noise ceiling.} Against the noise ceiling of the held-out LeBel story, the encoding models capture a median of 18 to 29\% of the explainable signal per subject, rising to 42 to 57\% in the best-predicted voxels, the bright areas of the accuracy maps in \Cref{fig:regions}.

\begin{figure}[!htbp]
\centering
\includegraphics[width=\linewidth]{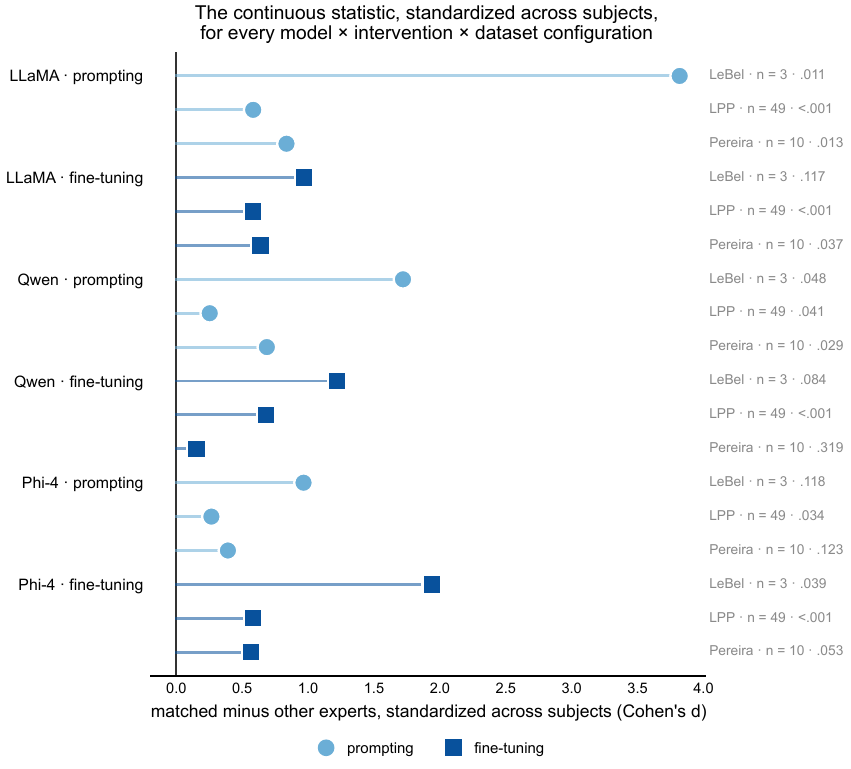}
\caption{\textbf{$\Delta r$ standardized across subjects for every configuration.} Cohen's d of the per-subject $\Delta r$, one point per model $\times$ intervention $\times$ dataset configuration.}
\label{fig:deltar_d}
\end{figure}

\paragraph{The continuous statistic across subjects.} \Cref{fig:deltar_d} shows $\Delta r$ standardized across subjects (Cohen's d) for all eighteen configurations.

\clearpage
\section{Additional Details on the Comparison of the Two Interventions}\label{app:interventions}

\begin{figure}[t]
\centering
\includegraphics[width=0.85\linewidth,height=0.86\textheight,keepaspectratio]{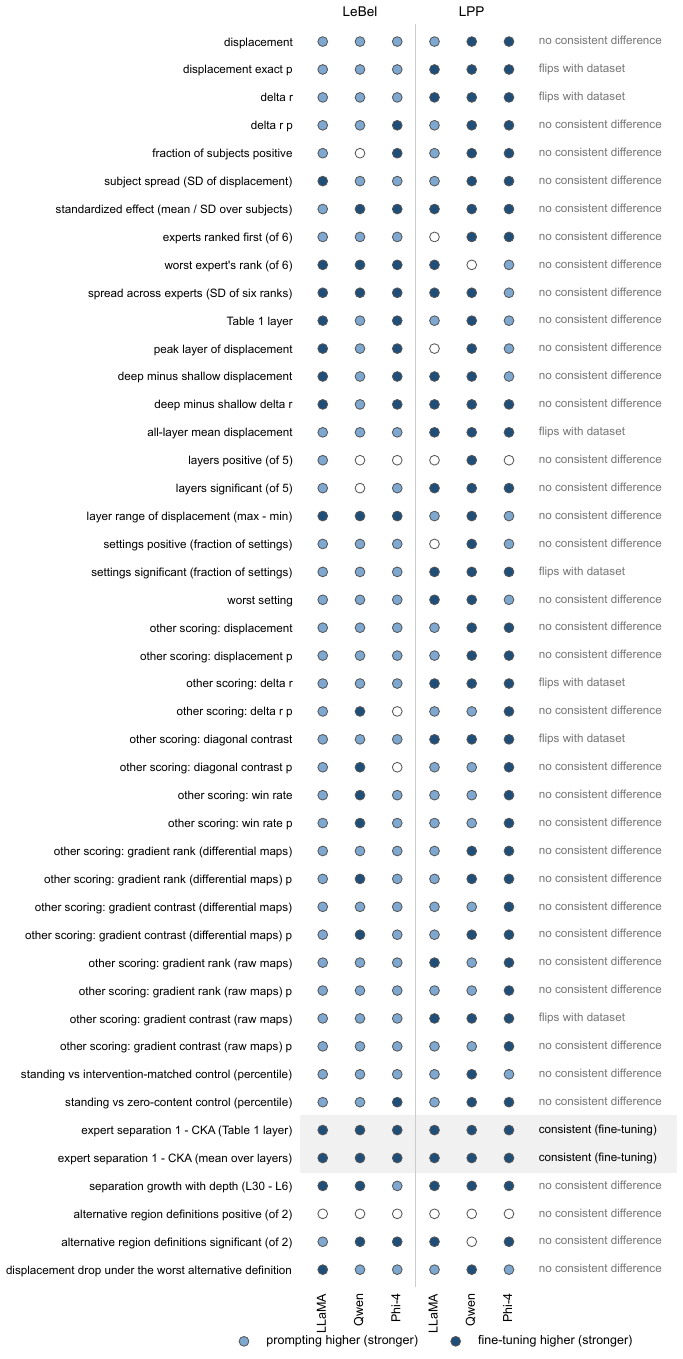}
\caption{\textbf{Every intervention contrast.} Which intervention has the larger (stronger) value on each of the 45 properties, in each of the six base model $\times$ dataset combinations; shaded rows are consistent across all six.}
\label{fig:route_ledger}
\end{figure}

\paragraph{The contrast ledger.} For each of 45 properties with a value for both interventions in every configuration, we took the six narrative configurations, LeBel and Le Petit Prince with three base models each, and recorded which intervention has the larger value (\Cref{fig:route_ledger}). A property separates the interventions consistently only when all three base models agree on both datasets in the same direction, which would happen by chance for about 1.4 of the 45. Representational separation does, measured at the reported layer and averaged over layers (6 of 6 each; shaded rows): fine-tuning's experts are further apart. Seven properties flip with the dataset, and they flip together: on these, prompting is the stronger intervention on LeBel for all three base models and fine-tuning on Le Petit Prince. On the remaining 36, among them consistency across subjects, the depth of the effect, and its robustness to layer, adapter rank, region definition and scoring statistic, the two interventions are comparable.

\paragraph{Representational separation.} Fine-tuning separates the six experts' stimulus representations increasingly with depth, reaching 1.4 to 6.8 times prompting's separation, while prompting keeps the experts close together at every layer (\Cref{fig:interventions}a); the surface-form control separates its members more than cognitive prompting does at every layer. Separation is one minus the mean pairwise linear CKA between the experts' representations of the same stimuli, which is 0 when two experts represent the stimuli identically up to rotation and scaling. The separation does not buy better alignment: the fine-tuned families whose prediction maps differ most align their experts least, within every dataset (\Cref{fig:interventions}b).

\begin{figure}[!htbp]
\centering
\includegraphics[width=0.6\linewidth]{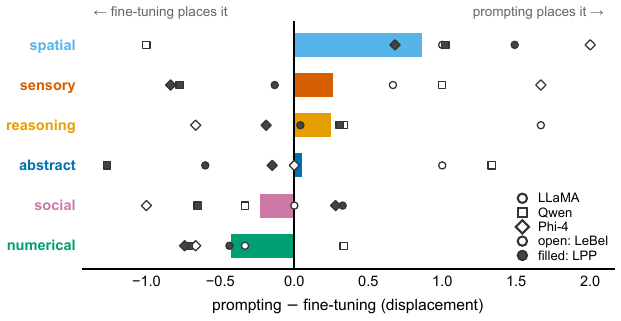}
\caption{\textbf{The intervention difference per expert.} Prompting minus fine-tuning own-region displacement per expert, pooled over the two narrative datasets and three base models (bar) with the six individual configurations (open LeBel, filled Le Petit Prince; circle LLaMA, square Qwen, diamond Phi-4).}
\label{fig:route_domains}
\end{figure}

\paragraph{Which experts each intervention aligns.} Prompting's spatial expert is above chance in its own region in all six narrative configurations and fine-tuning's in three, with a mean of 0.00. The difference holds within subject on Le Petit Prince for each base model (t(48) = 5.0, 2.6 and 2.4; pooled 5.8) and at every layer and adapter rank, and there the cognitive instruction aligns the spatial expert where a random instruction does not (+0.95 over the random-prompt control, t(48) = 4.65, $p$ \textless{} .0001); the surface-form adapter does not align it either (+0.11). The numerical expert shows the reverse: fine-tuning's is above chance in all six narrative configurations (mean +1.39, against +0.96 for prompting) and outperforms prompting's within subject on Le Petit Prince at every layer (t(48) between 2.3 and 5.2, all p \textless{} .03); the social expert is aligned about equally by both (means +1.17 and +0.94, no layer with p \textless{} .1). The remaining experts differ by dataset rather than by intervention. Which experts align is something the brain data reveal and behavior does not. Although base models whose experts are more specialized align better overall (\Cref{sec:localization}), within each base model and intervention the rank correlation across the six experts between behavioral gain and brain alignment lies between $-$0.6 and +0.5 with every $p$ \textgreater{} .2, and the spatial expert is the second or third most specialized by lexicon share and perplexity in every base model. \Cref{fig:route_domains} shows the intervention difference for each expert.

\clearpage
\section{Additional Details on the Control Analyses}\label{app:controls}

\begin{figure}[t]
\centering
\includegraphics[width=\linewidth]{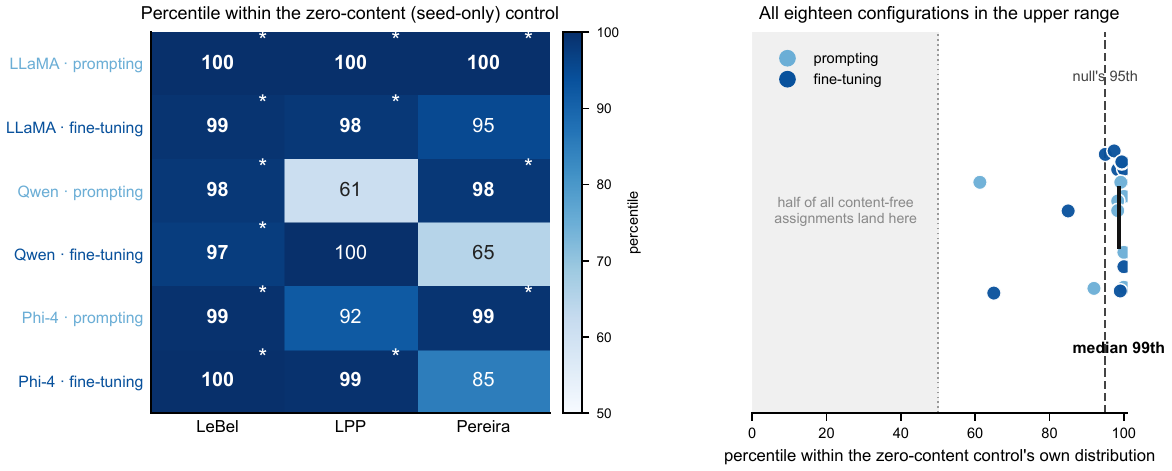}
\caption{\textbf{Where every configuration falls inside the null.} \emph{(a)} Each of the eighteen configurations of \Cref{tab:main} as its percentile within the seed-only control family's assignment distribution; asterisks mark individually significant configurations. \emph{(b)} The same eighteen values on one axis; the shaded region is the lower half of the distribution, where a random assignment falls with probability one half.}
\label{fig:null_standing}
\end{figure}

\paragraph{Every configuration against the null.} \Cref{fig:null_standing} places each of the eighteen configurations of \Cref{tab:main} within the seed-only family's distribution of displacement over all assignments of its members to the regions. A random assignment would fall anywhere in that distribution with equal probability, so its expected standing is the 50th percentile. Every configuration lands in the upper range: the lowest at the 61st percentile, the median at the 99th, and fourteen of eighteen beyond the 95th. Averaged over the eighteen, the expert families stand at the 94th percentile of the seed-only null and at the 89th of the intervention-matched controls, against an exact joint null that relabels all of a dataset's configurations together ($p$ \textless{} $10^{-6}$ for both means). These standings are read at each configuration's reported setting; given the same sweep as the expert families, the intervention-matched controls reach a mean displacement of +0.23 against the experts' +0.55 (\Cref{app:inference}).

\begin{figure}[t]
\centering
\includegraphics[width=\linewidth]{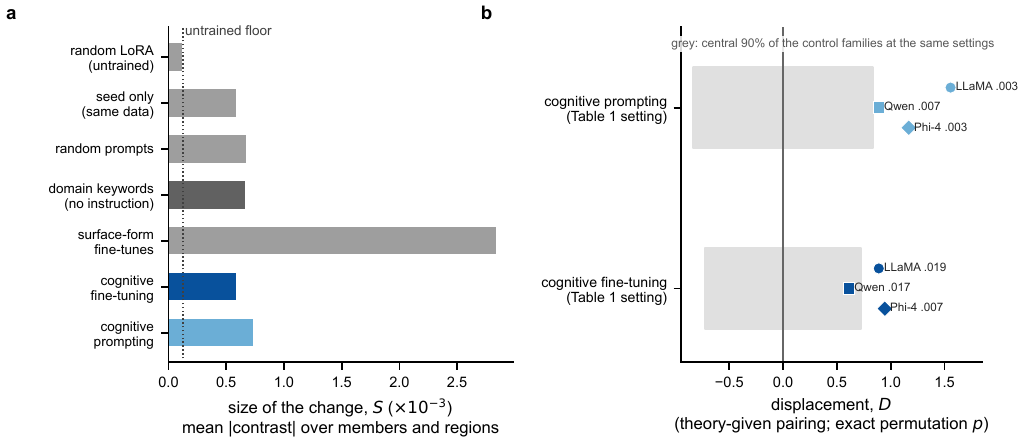}
\caption{\textbf{The size of a change and the place it lands.} \emph{(a)} The size of the change, $S$: the members' contrasts in absolute value, averaged over members, regions and subjects, with no ranking and no pairing to regions. \emph{(b)} Displacement, $D$, under the theory-given pairing with its exact permutation $p$, for the expert families at their \Cref{tab:main} settings (circle LLaMA, square Qwen, diamond Phi-4), against the central 90\% of the control families at the same settings (grey).}
\label{fig:size_vs_place}
\end{figure}

\paragraph{The size of a change and where it lands.} Displacement and the size of a change are two different summaries of the same contrasts $c_{e,i}$ (\Cref{app:brain}). Displacement uses only their order within each region, and it needs a pairing $\pi$ of regions to members. An expert family is scored at the pairing given by theory. A control family has no such pairing, so $D$ is computed under every one of the $K!$ pairings; these values are its null distribution, which averages exactly 0, and an expert family's percentile is its $D$ located within that distribution. The size of the change uses only the magnitudes of the contrasts and needs no ranking and no pairing:
\[
S \;=\; \frac{1}{K^2}\sum_{e=1}^{K}\sum_{i=1}^{K} \left| c_{e,i} \right|,
\]
averaged over subjects and layers. Multiplying every contrast by a constant leaves $D$ unchanged and scales $S$, so the two can move independently (\Cref{fig:size_vs_place}). The surface-form fine-tunes have the largest $S$ of any family, nearly four times that of the largest expert family, and the untrained random-LoRA adapters almost none, between a fifth and a twenty-third of any trained or prompted family; it is the expert families, at intermediate $S$, whose theory-given pairing stands out in $D$. The seed-only fine-tunes have an ordinary $S$ and a null distribution as wide as the surface-form family's, so the width of a null reflects the sampling noise of a six-way ranking rather than the size of the change. A large change is therefore neither necessary nor sufficient for alignment; what matters is whether it follows the cognitive domains.

\paragraph{The keyword family.} \Cref{tab:keyword} scores the keyword family, bare domain vocabulary with no instruction, on all three base models and all three datasets, read at the same setting as the cognitive prompting family beside it. The instruction aligns more strongly than the bare vocabulary in seven of the nine comparisons, so naming a domain's words is not enough, and instructing the model to act as an expert in the domain is needed to align it with the corresponding brain system.

\begin{table}[t]
\caption{\textbf{Instruction against bare vocabulary in every base model $\times$ dataset comparison.} Own-region displacement of the cognitive prompting family (instruction) and of the keyword family (vocabulary), both read at the expert family's \Cref{tab:main} setting, with each family's exact relabeling $p$ (bold: displacement with $p$ below .05).}
\label{tab:keyword}
\centering
\small
\setlength{\tabcolsep}{4pt}
\begin{tabular}{llrrrrr}
\toprule
Dataset & Model & Instruction & $p$ & Vocabulary & $p$ & Instruction ahead \\
\midrule
LeBel & LLaMA & \textbf{+1.56} & .003 & +0.22 & .342 & yes \\
LeBel & Qwen & \textbf{+0.89} & .007 & \textbf{+1.22} & .004 & no \\
LeBel & Phi-4 & \textbf{+1.17} & .003 & $-$0.17 & .654 & yes \\
LPP & LLaMA & \textbf{+0.42} & .017 & +0.12 & .242 & yes \\
LPP & Qwen & +0.07 & .321 & +0.18 & .151 & no \\
LPP & Phi-4 & +0.17 & .214 & +0.16 & .226 & yes \\
Pereira & LLaMA & \textbf{+0.54} & .017 & +0.16 & .225 & yes \\
Pereira & Qwen & \textbf{+0.43} & .042 & +0.15 & .167 & yes \\
Pereira & Phi-4 & \textbf{+0.42} & .033 & +0.32 & .075 & yes \\
\bottomrule
\end{tabular}
\end{table}

\begin{figure}[!htbp]
\centering
\includegraphics[width=0.9\linewidth]{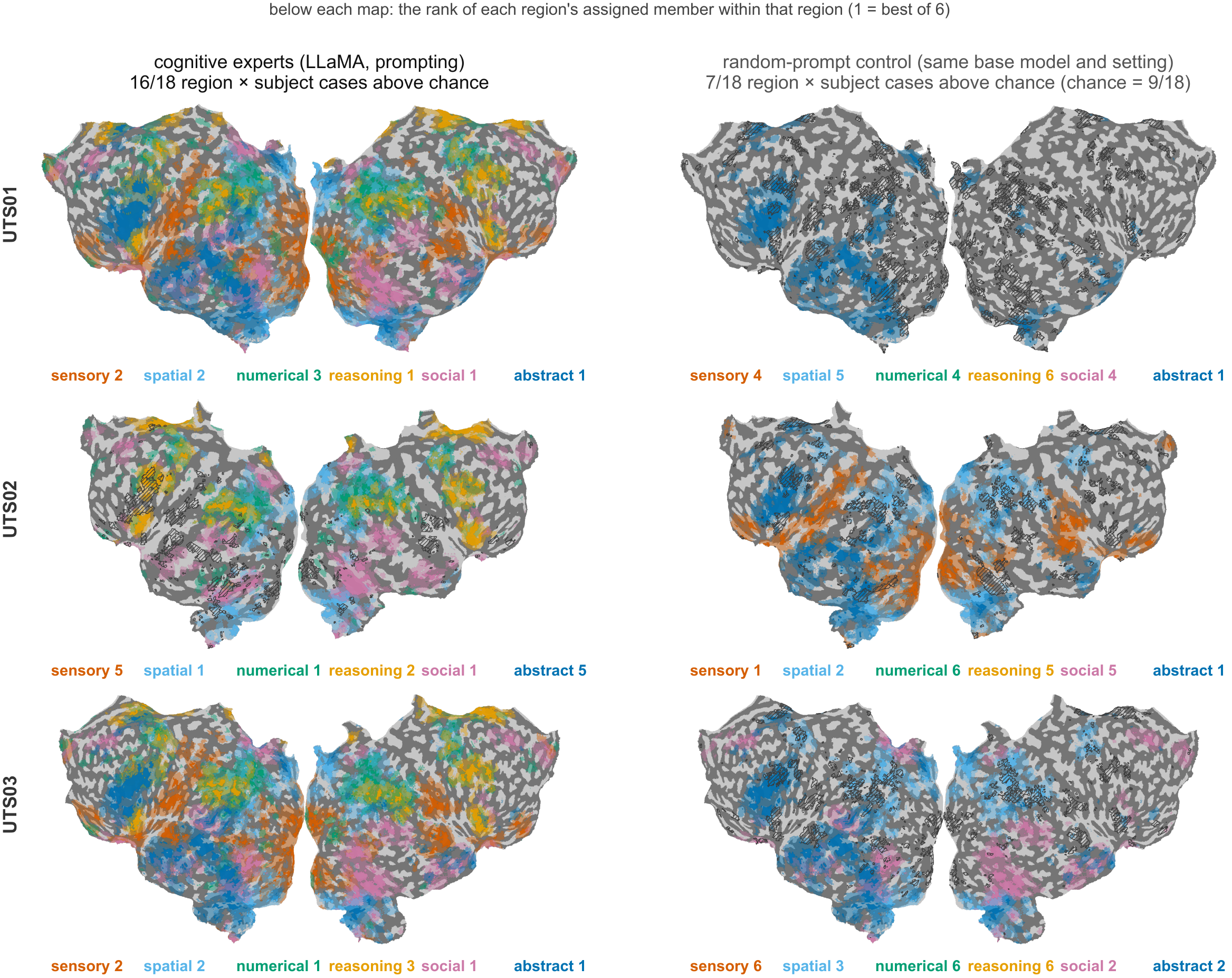}
\caption{\textbf{The same cortical map for cognitive experts and for a control.} Left, the running example (LLaMA, prompting) on the conventions of \Cref{fig:lead}; right, the random-prompt control at the same setting under one arbitrary assignment. A region is filled when its assigned expert ranks above chance inside it and hatched when it does not.}
\label{fig:map_vs_control}
\end{figure}

\paragraph{The cortical map of a control.} \Cref{fig:map_vs_control} draws the running example's map beside the same map for the random-prompt control, six non-cognitive system prompts on the same base model at the same setting, under one arbitrary assignment of its members to the regions. The cognitive experts rank above chance in 16 of 18 region $\times$ subject cases; the control reaches 7, close to the 9 expected by chance. 

\clearpage
\section{Additional Details About Experimental Environment}\label{app:environment}

All fine-tuning, feature extraction and encoding fits ran on two GPU nodes of a shared cluster, one with eight NVIDIA RTX A6000 cards and one with four NVIDIA H200 NVL cards, with every job on a single GPU. The software stack is Python 3.10.12 with PyTorch 2.10 and Transformers 4.57 with PEFT 0.18 for fine-tuning, PyTorch 2.12 and Transformers 5.10 for feature extraction and encoding, NumPy 2.2, SciPy 1.15 and scikit-learn 1.7, nibabel 5.4 and nilearn 0.13 for the imaging data, pycortex 1.3 for the cortical surfaces, and the FSL linear registration tools \citep{jenkinson2002improved,smith2004advances} for projecting the region masks with each dataset's own released transforms; we ran no new preprocessing on any dataset. Fine-tuning one adapter takes a median of 3.5 hours on one GPU (the middle half of runs between 0.9 and 5.5 hours), the spread coming from the domains' unequal training sets and the base models' different sizes. On the narrative datasets, extracting one expert's representations of the stimuli takes about 17 minutes, and fitting the encoding model for one expert at one layer in one subject about 3 minutes; the sentence dataset is far cheaper, a complete family finishing in minutes.

\end{document}

%% file: math_commands.tex
\usepackage{amsmath,amsfonts,bm}

\def\eqref#1{equation~\ref{#1}}

\def\1{\bm{1}}

\DeclareMathAlphabet{\mathsfit}{\encodingdefault}{\sfdefault}{m}{sl}
\SetMathAlphabet{\mathsfit}{bold}{\encodingdefault}{\sfdefault}{bx}{n}

